\documentclass{article}

\usepackage[preprint]{neurips_2025}

\usepackage[utf8]{inputenc} 
\usepackage[T1]{fontenc}    
\usepackage{hyperref}       
\usepackage{url}            
\usepackage{booktabs}       
\usepackage{amsfonts}       
\usepackage{nicefrac}       
\usepackage{microtype}      
\usepackage{xcolor}         

\usepackage{amsmath}
\usepackage{amssymb}
\usepackage{mathtools}
\usepackage{amsthm}

\usepackage{bbding}
\usepackage{caption}
\usepackage{subcaption}
\usepackage{svg}
\usepackage{multirow}  

\title{ASRC-SNN: Adaptive Skip Recurrent Connection Spiking Neural Network}

\author{%
Shang Xu\textsuperscript{1} \quad
Jiayu Zhang\textsuperscript{1} \quad
Ziming Wang\textsuperscript{1} \quad
Runhao Jiang\textsuperscript{1} \quad
Rui Yan\textsuperscript{2} \quad
Huajin Tang\textsuperscript{1} \thanks{Correspondence: htang@zju.edu.cn} \\
\textsuperscript{1}College of Computer Science, Zhejiang University \\
\textsuperscript{2}College of Computer Science, Zhejiang University of Technology  \\
}

\begin{document}

\maketitle

\begin{abstract}
  In recent years, Recurrent Spiking Neural Networks (RSNNs) have 
shown promising potential in long-term temporal modeling. Many studies focus on improving neuron models and also integrate recurrent structures, leveraging their synergistic effects to improve the long-term temporal modeling capabilities of Spiking Neural Networks (SNNs). However, these studies often place an excessive emphasis on the role of neurons, overlooking the importance of analyzing neurons and recurrent structures as an integrated framework. In this work, we consider neurons and recurrent structures as an integrated system and conduct a systematic analysis of gradient propagation along the temporal dimension, revealing a challenging gradient vanishing problem. To address this issue, we propose the Skip Recurrent Connection (SRC) as a replacement for the vanilla recurrent structure, effectively mitigating the gradient vanishing problem and enhancing long-term temporal modeling performance. Additionally, 
we propose the Adaptive Skip Recurrent Connection (ASRC), a method that can learn the skip span of skip recurrent connection in each layer of the network. Experiments show that replacing the vanilla recurrent structure in RSNN with SRC significantly improves the model's performance on temporal benchmark datasets. Moreover, ASRC-SNN outperforms SRC-SNN in terms of temporal modeling capabilities and robustness. Our code is available at \url{https://github.com/dgxdn/ASRC-SNN.git}.
\end{abstract}

\section{Introduction}
\label{introduction}
In artificial neural networks (ANNs), activation values are continuous, and computations involve numerous computationally expensive multiply-accumulation (MAC) operations. 
In contrast, in spiking neural networks (SNNs), the activation is represented by binary spike signals, where most MAC operations can be replaced by energy-efficient accumulation (AC) operations. 
Moreover, the sparse generation of spikes \cite{roy2019sparse0, nunes2022sparse} in SNN further reduces the number of AC operations. As a result, SNNs have a significant advantage in energy consumption compared to ANNs. This theoretical advantage has been validated in practice with SNNs deployed on neuromorphic hardware that demonstrate fast inference and low power consumption \cite{akopyan2015chip, davies2018chip, pei2019chip}.  \par
Leaky Integrate-and-Fire (LIF) neuron model \citep{gerstner2002book_lif}, due to their computational efficiency and similarity to biological neurons, have become the most widely used in SNN. Furthermore, the favorable temporal dynamics of LIF neurons makes LIF neuron-based SNNs well suited for handling temporal tasks. Nowadays, for static image classification tasks, replicating the image multiple times along the temporal dimension to introduce simple temporal features has enabled SNNs to achieve performance comparable to that of ANNs \cite{ding2021conversion, zhou2023spikformer, yao2024sdriven, zhou2024qkformer}. 
For tasks that align with the event-driven paradigm and incorporate intrinsic temporal features, such as neuromorphic image datasets like CIFAR-10-DVS \cite{li2017cifar10} and DVS-128-Gesture \cite{amir2017dvs128}, which are captured using Dynamic Vision Sensors (DVS) \cite{lenero2011dvs}, SNNs have demonstrated exceptional performance \cite{deng2023surrogate, ma2023NSNN, wang2023ASGL, huang2024clif}. \par
The tasks mentioned above are characterized by short time steps and simple temporal dependencies. 
For complex tasks that require the establishment of long-term temporal dependencies, such as speech recognition and sequence recognition, SNNs that rely solely on LIF neurons to capture temporal relationships are generally less competitive in terms of performance compared to ANNs \cite{yin2021alif}. To enhance the competitiveness of SNNs, a common approach is to introduce the recurrent structure \cite{elman1990vanillarnn} and improve the neuron model based on simple SNNs \cite{yin2021alif, bittar2022radlif, zhang2024tc-lif, baronig2024advancing}. In analyzing the temporal modeling capabilities of these improved models, these studies primarily focus on the function of neurons, while treating the recurrent connection as an additional mechanism aimed at enhancing the model performance. We believe that both the recurrent connection and the neuron play a synergistic role in capturing temporal dependencies. This work is the first to treat both components as a unified system for gradient analysis along the temporal dimension. This reveals that vanilla recurrent spiking neural networks (RSNNs) based on LIF neurons suffer from vanishing and exploding gradients when gradients propagate along the temporal dimension. At the same time, we provide the corresponding solutions to address these issues. Specifically, the exploding gradient problem, which arises from the recurrent structure, can be mitigated by orthogonal initialization \cite{henaff2016orthogonal}; the challenging vanishing gradient problem, which arises from both the LIF neurons and the recurrent structure, can be alleviated by replacing the vanilla recurrent connection with the skip recurrent connection (SRC). Experiments show that SRC-SNN significantly outperforms vanilla RSNN in long-term temporal tasks. \par
Furthermore, We identified two limitations of SRC: The uniform skip connection span across layers in SRC-SNN constrains the network's temporal modeling ability; the hyparameter tuning process is complex. To address these limitations, we propose the adaptive skip recurrent connection (ASRC). For each layer in the ASRC-SNN, we introduce a temperature-scaled Softmax kernel. This Softmax kernel enables competition among multiple skip connections of varying temporal spans, while the temperature parameter gradually decreasing to intensify this competition. Specifically, initially, the kernel assigns equal weights to multiple skip connections; as training progresses, the temperature parameter decreases, and the Softmax kernel progressively concentrates the weights on the most relevant skip connection. Experiments show that ASRC-SNN demonstrates superior long-term temporal modeling capabilities and robustness compared to SRC-SNN. 
\par
The main contributions of this paper are as follows:

1) We are the first to unify recurrent structures and neuron models within an integrated framework to investigate gradient behavior along the temporal dimension, revealing a challenging gradient vanishing problem.

2) To mitigate this problem, we propose replacing the standard recurrent structure in RSNNs with Skip Recurrent Connections (SRC), explicitly considering the interaction between the recurrent dynamics and neuronal properties. Empirical results demonstrate that this architectural change significantly improves performance on long-term sequence tasks.

3) While SRC offers notable improvements, we identify two key limitations that hinder its effectiveness. To address these, we introduce Adaptive Skip Recurrent Connections (ASRC), which enhance long-term temporal modeling and improve robustness. Across four temporal benchmarks, ASRC consistently outperforms SRC and surpasses most prior SNN models.

4) ASRC dynamically adjusts skip spans, providing a novel perspective for mitigating temporal gradient vanishing. The essence of ASRC lies in learning a discrete position along the temporal dimension, with the potential to extend this method to learning a discrete position in both time and space.

\section{Related Work}

\subsection{Long-term temporal modeling in RSNNs}
ALIF \cite{yin2021alif} extends the LIF neuron model by incorporating a dynamic threshold mechanism, thereby enhancing their ability to model temporal sequences. \cite{bittar2022radlif, baronig2024advancing} argues that the dynamics of LIF neurons is relatively simple and proposes the introduction of a second time-varying variable to model the oscillatory behavior. \cite{zhang2024tc-lif} have proposed a novel biologically inspired two-compartment leaky integrative-and-fire (TC-LIF) spiking neuron model, which integrates specifically designed somatic and dendritic compartments aimed at improving the learning of long-term temporal dependencies. To further improve model performance, these enhanced neuron models incorporate the recurrent structure \cite{elman1990vanillarnn}. Inspired by the success of gated recurrent units (GRUs) in ANNs, 
\cite{dampfhoffer2022spikingGRU} have proposed the spiking GRU model. \cite{wang2024STC} introduces a novel spatial-temporal circuit (STC) model that incorporates two learnable adaptive pathways, improving the temporal memory capabilities of spiking neurons.
\subsection{Long-term temporal modeling in other SNNs}
By reformulating the neuronal dynamics without reset into a general mathematical form, \cite{fang2024PSN} introduces a series of parallel spiking neuron (PSN) models, which transform the membrane potential charging dynamics into a learnable decay matrix. The parallel multi-compartment spiking neuron (PMSN) \cite{chen2024pmsn} mimics biological neurons by incorporating multiple interacting substructures, enabling effective representation of temporal information across diverse timescales. Both of these studies propose corresponding parallelization methods to enhance computational efficiency. The balanced resonate-and-fire neuron (BRFN) \cite{higuchi2024bhrf} builds upon the resonate-and-fire neuron by incorporating a dynamic threshold mechanism to simulate the refractory period, thus effectively maintaining the stability of the oscillatory process. DCLS-delays \cite{hammouamri2024dcls} demonstrates strong performance in speech datasets by learning synaptic delays. Additionally, some studies have drawn inspiration from successful architectures in ANNs and effectively applied them to SNNs. For example, \cite{yao2021TA-SNN} introduces the attention mechanism, \cite{sadovsky2023snn-cnn} utilizes convolutional neural networks (CNNs), \cite{liu2024lmuformer} incorporates the legendre memory unit (LMU) and \cite{stan2024spikeSSM0, shen2024spikingssms} explore state space models (SSMs).

\section{Methods}
\label{Methods}
In this section, we first introduce the LIF neuron and then present the basic paradigm of RSNNs based on LIF neurons, with a focus on analyzing gradient propagation over the time dimension. Through this analysis, we identify the issues of vanishing and exploding gradients. For the vanishing gradient issue, we propose replacing the vanilla recurrent structure with the SRC structure. Finally, we discuss the limitations of the SRC-SNN model and introduce the ASRC-SNN model to overcome these limitations.
\subsection{Preliminary}

\subsubsection{LIF Neuron Model}
The LIF neuron is the most commonly used neuron model in the field of SNNs, known for its simplicity and computational efficiency while still capturing key aspects of neuronal dynamics. Its mathematical expressions are given by the following equations:
\begin{equation} \label{eq:1}
U^{l}[t]=\alpha U^{l}[t-1]+I^{l}[t]
\end{equation}
\begin{equation} \label{eq:2}
S^{l}[t]=H(U^{l}[t]-V_{th})
\end{equation}
\begin{equation} \label{eq:3}
U^{l}[t]=U^{l}[t]-V_{th}S^{l}[t]
\end{equation}
Here, $U^{l}[t]$ and $I^{l}[t]$ represent the membrane potential and the input current of the neuron in the $l$-th layer at time step $t$. $\alpha$ is a decay factor that controls membrane potential leakage and ranges from 0 to 1. Eq.~\eqref{eq:1} describes how the membrane potential evolves over time by integrating the previous membrane potential with the input current, while the decay factor reflects the natural leakage of the potential. $H(\cdot)$ is Heaviside function. $S^{l}[t]$ represent the spike state of the $l$-th layer neuron at time step $t$, where $S^{l}[t]=1$ if the neuron fires a spike and $S^{l}[t]=0$ otherwise. Eq.~\eqref{eq:2} represents that if the membrane potential $U^{l}[t]$ exceeds the threshold $V_{th}$, the neuron fires; otherwise, no spike occurs. Eq.~\eqref{eq:3} describes the reset process of the membrane potential after a spike is fired. This soft reset mechanism is designed to preserve more information, as suggested by \cite{rueckauer2017softreset, han2020ssoftreset, huang2024clif}. \par
\subsubsection{Paradigm of LIF-based RSNN}
The SNN without vanilla recurrent connections has $I^{l}[t]=W_{1}^{l}S^{l-1}[t]$, while the SNN with vanilla recurrent connections has $I^{l}[t]=W_{1}^{l}S^{l-1}[t]+W_{2}^{l}S^{l}[t-1]$. Here $W_{1}^{l}$ and $W_{2}^{l}$ represent the parameters of the feedforward connections and recurrent connections, respectively, in the $l$-th layer. This paper focuses on the mechanisms of RSNNs. By substituting $I^{l}[t]=W_{1}^{l}S^{l-1}[t]+W_{2}^{l}S^{l}[t-1]$ into Eq.~\eqref{eq:1} and integrating Equations Eq.~\eqref{eq:1} and Eq.~\eqref{eq:3}, the membrane potential update equation for the neurons in the $l$-th layer of the RSNN can be derived:
\begin{equation} \label{eq:4}
U^{l}[t]=\alpha (U^{l}[t-1]-V_{th}S^{l}[t-1])+W_{1}^{l}S^{l-1}[t]+W_{2}^{l}S^{l}[t-1]
\end{equation}
In a vanilla LIF-based RSNN, the $l$-th layer can be described by Eq.~\eqref{eq:4} and Eq.~\eqref{eq:2}, with $S^0$ considered as the network's input.
\subsection{Temporal Gradient Analysis of LIF-based RSNN}
\label{sec:theroy}
\cite{yin2021alif, bittar2022radlif, baronig2024advancing, zhang2024tc-lif} treat the recurrent connection as an additional mechanism aimed at improving the model's performance. In contrast, we consider the recurrent structure and neurons as working synergistically, analyzing them within a unified framework. Considering the propagation of gradients across adjacent time steps, we have:
\begin{equation} \label{eq:5}
\frac{\partial U^{l}[t+1]}{\partial U^{l}[t]}=\alpha+(W_{2}^{l}-\alpha V_{th})\frac{\partial S^{l}[t]}{\partial U^{l}[t]}
\end{equation}
The Heaviside function is non-differentiable, and a common approach is to use a surrogate gradient function to approximate its derivative \cite{neftci2019surrogate}. Similar to \cite{deng2022TET, zhang2024tc-lif}, we use the triangle function as the surrogate gradient function:
\begin{equation} \label{eq:6}
\frac{\partial S^{l}[t]}{\partial U^{l}[t]} \approx \mathbb{H} (U^{l}[t]) = \frac{1}{\gamma^{2}}\max(0, \gamma - \lvert U^{l}[t]-V_{th} \rvert)
\end{equation}
where $\gamma$ represents the constraint factor that governs the range of samples required to activate the gradient. In this work, we set $\gamma=V_{th}$. 
Considering the propagation of the gradient over a longer time span, we have:
\begin{equation} \label{eq:7}
\frac{\partial U^{l}[t+k]}{\partial U^{l}[t]} =
\frac{\partial U^{l}[t+k]}{\partial U^{l}[t+k-1]}
\frac{\partial U^{l}[t+k-1]}{\partial U^{l}[t+k-2]}
\dots
\frac{\partial U^{l}[t+1]}{\partial U^{l}[t]} 
=
\prod_{t'=0}^{k-1}
(
\alpha+(W_{2}^{l}-\alpha V_{th})\mathbb{H} (U^{l}[t+t'])
)
\end{equation}

Considering the extreme case where
$\mathbb{H} (U^{l}[t+t'])$ is always equal to $\frac{1}{V_{th}}$, we have
$
\frac{\partial U^{l}[t+k]}{\partial U^{l}[t]}=(\frac{W_2^l}{V_{th}})^{k}
$.
When the value of $k$ is sufficiently large, and $\lvert W_2^l\rvert > V_{th}$, the gradient explosion problem occurs in the temporal dimension. The gradient explosion problem caused by the recurrent structure. This issue can be addressed using recurrent neural networks (RNNs) techniques \cite{pascanu2013gradclip, henaff2016orthogonal}. To mitigate this, we employ orthogonal initialization \cite{henaff2016orthogonal}, a simple and effective approach. \par
When  $\lvert W_2^l \rvert\leq V_{th}$, combining Equations  Eq.~\eqref{eq:6} and  Eq.~\eqref{eq:7}, we have:
\begin{equation} \label{eq:8}
\lvert \frac{\partial U^{l}[t+k]}{\partial U^{l}[t]} \rvert
\leq
\max(\alpha ^k, \lvert \frac{W_2^l}{V_{th}} \rvert ^k)
\leq
1
\end{equation}
 
When $k$ is large enough, the problem of gradient vanishing in the temporal dimension can be avoided only if both inequality signs in Eq.~\eqref{eq:8} become equalities. To meet this condition, it is essential that $\lvert W_2^l \rvert=V_{th}$ and $\mathbb{H} (U^{l}[t+t'])=\frac{1}{V_{th}}$  for $0\leq t' <k$. However, this imposes a highly stringent constraint. More importantly, if the second condition is met, the neuron will fire continuously during the entire period, making it impossible to distinguish different states along the temporal dimension. As a result, RSNNs are prone to vanishing gradients in the temporal dimension, limiting their ability to capture long-term dependencies. To mitigate the gradient vanishing problem, improvements can be made from either the LIF neuron or the recurrent structure perspective while maintaining their coordination. This work primarily focuses on optimizing the recurrent structure.

\begin{figure*}[t!]
    \centering
    \includegraphics[width=0.8\textwidth]{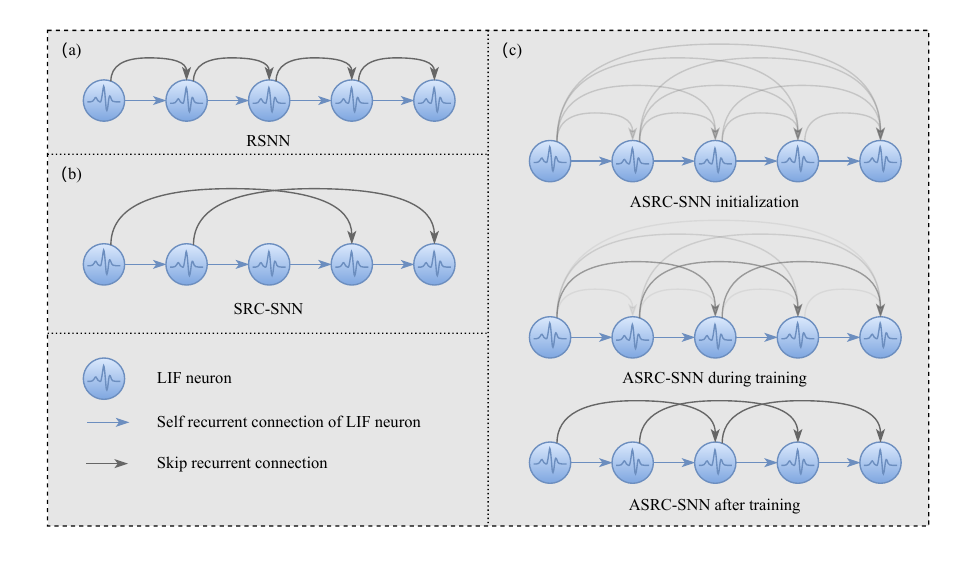}
    \caption{This figure demonstrates the flow of temporal information within the LIF neurons of the vanilla RSNN, SRC-SNN, and ASRC-SNN models. (a) illustrates that in RSNN, recurrent connections are restricted to adjacent time steps. (b) illustrates that in SRC-SNN, recurrent connections can span multiple time steps. (c) illustrates the dynamic evolution of skip recurrent connections in ASRC-SNN from the start to the end of training. ASRC-SNN initialization: At the beginning of training, each LIF neuron in ASRC-SNN is connected to $T_{\lambda}$ skip recurrent connections, with their weights initialized to $\frac{1}{T_{\lambda}}$. ASRC-SNN during training: During training, the weight distribution of the $T_{\lambda}$ skip recurrent connections becomes more concentrated. ASRC-SNN after training: After training, the weights of the $T_{\lambda}$ skip recurrent connections converge onto a single skip recurrent connection.}
    \label{fig:fig_all}  
\end{figure*}

\subsection{Skip Recurrent Connection}
Skip recurrent connections (SRC) can alleviate the vanishing gradient problem by introducing direct pathways between temporal steps. We propose SRC-SNN by replacing the vanilla recurrent structure in RSNN with SRC. In SRC-SNN, The membrane potential update equation for the $l$-th layer neuron is given as follows:
\begin{equation} \label{eq:9}
U^{l}[t]=\underbrace{ \alpha (U^{l}[t-1]-V_{th}S^{l}[t-1]) }_{\text{self-connections of LIF neurons}}+\underbrace{W_{1}^{l}S^{l-1}[t]}_{\text{feedforward connections}} 
+\underbrace{W_{2}^{l}S^{l}[t-\lambda]}_{\text{skip recurrent connections}}
\end{equation}

Here, $\lambda$ represents the skip coefficient, which is typically greater than 1 (Figure.\ref{fig:fig_all}b). When $\lambda = 1$, SRC-SNN degenerates into vanilla RSNN (Figure.\ref{fig:fig_all}a). Furthermore, we found that \cite{zhang2016skippedrnn} systematically analyzed and validated the effectiveness of SRC in modeling long-term dependencies. It is worth noting that, unlike \cite{zhang2016skippedrnn}, where adjacent time steps are connected through the vanilla recurrent structure, in SRC-SNN, the connections between adjacent time steps are inherently supported by the intrinsic self-connections of LIF neurons. \par
\textbf{The limitations of SRC} Individually tuning the hyperparameters of the skip coefficients for each layer in SRC-SNN results in an exponential growth in the number of hyperparameter optimization experiments, making this approach impractical. Consequently, the skip coefficients across the layers in SRC-SNN are set to be identical, which, however, limits the temporal modeling capability of the SRC. Moreover, the optimal setting of the skip coefficients in the SRC-SNN models often differs on different datasets, and in some cases the model performance is highly sensitive to changes in the skip coefficient (see Section \ref{sec:Ablation_SRC}). Therefore, the hyperparameter $\lambda$ tuning process in SRC-SNN is complex.
\subsection{Adaptive Skip Recurrent Connection}
\subsubsection{The Design of ASRC}
To address the limitations of SRC, we propose an improved approach termed adaptive skip recurrent connection (ASRC), which can learn the span of skip connection. This approach is inspired by the asymptotic behavior of the Softmax function in the low-temperature regime \cite{guo2017calibration}. Specifically, when the Softmax function is parameterized with a temperature $\tau > 0$
, its form is given by:
\begin{equation} \label{eq:10}
\text{Softmax}_\tau(x_i) = \frac{\exp(x_i / \tau)}{\sum_{j} \exp(x_j / \tau)}
\end{equation}
where \( x = [x_1, x_2, \dots, x_n] \) denotes the input vector and \( x_i \) represents the \( i \)-th element. As \( \tau \to 0 \) (the low-temperature limit), the Softmax function exhibits the following asymptotic behavior:
\begin{equation} \label{eq:11}
\lim_{\tau \to 0} \text{Softmax}_\tau(x_i) =
\begin{cases}
1, & \text{if } i = \arg\max_j x_j \\
0, & \text{otherwise}
\end{cases}
\end{equation}
In this limit, the Softmax function converges to a Hardmax operation, where the output corresponds to a one-hot encoding of the position of the maximum value in the input vector.\par
In the ASRC-SNN, the membrane potential update equation for the $l$-th layer neurons is as follows:
\begin{equation} \label{eq:12}
U^{l}[t]=\alpha (U^{l}[t-1]-V_{th}S^{l}[t-1])
+W_{1}^{l}S^{l-1}[t]+W_{2}^{l}\sum_{t'=1}^{T_\lambda} p^l[t']S^{l}[t-t']
\end{equation}
Here, $p^l$ represents the weights of multiple skip connections with varying temporal spans, which are computed using a Softmax kernel function with a temperature parameter $\tau$. $T_{\lambda}$ represents the length of the Softmax kernel, defining the maximum time span considered in the ASRC model, i.e., the longest time span that the skip recurrent connections can extend across. Specifically, $p^l$ regulates the influence of states from the past $T_{\lambda}$ time steps, distributing weights for skip connections and determining their contribution to the current neuron state update. The mathematical expression is as follows:
\begin{equation} \label{eq:13}
\forall t \in \{1, \dots, T_{\lambda}\},\,
p^l[t] = 
\frac{\exp(w^l[t] / \tau)}{\sum_{t'=1}^{T_\lambda} \exp(w^l[t'] / \tau)}
\end{equation}
Here, $w^l$ is a vector at the $l$-th layer containing $T_{\lambda}$ trainable parameters, initialized to zero. The temperature parameter $\tau$ is a non-trainable constant initialized to 1 and decreases after each epoch according to an exponential decay strategy. In our experiments, the exponential decay factor is set to 0.96. Indeed, during the model testing phase, we use Eq.~\eqref{eq:11} to compute $p^{l}$. \par
\subsubsection{Dynamic Analysis}
In the early stages of training, the distribution of the output of the Softmax kernel function exhibits high smoothness, enabling the simultaneous activation of multiple skip recurrent connections with varying temporal spans. During this phase, the model leverages this smooth selection mechanism to thoroughly explore the dependencies between the current state and multiple historical time steps, facilitating comprehensive temporal modeling. As training progresses, the temperature parameter gradually decreases, leading to a sharper distribution of weights in the Softmax kernel. Eventually, as the temperature approaches zero, the Softmax kernel function gradually converges to the Hardmax operation. At this stage, the model independently selects the most relevant time step for the skip connection at each layer, based on the current state of that layer. This hardening process (Figure.\ref{fig:fig_all}c) allows the model to focus more precisely on critical temporal dependencies. \par
\textbf{Enhanced Temporal Modeling Capacity} By dynamically adjusting the weights of the skip connections with varying temporal spans, ASRC can more accurately select the relevant time step for the skip connection. This flexible adjustment allows ASRC-SNN to better accommodate the varying temporal dependencies required by different layers, compared to the fixed setting of identical coefficients across layers in SRC-SNN. As a result, ASRC exhibits enhanced temporal modeling capabilities. \par
\textbf{Enhanced Robustness} 
ASRC-SNN is capable of adaptively adjusting the skip recurrent connections based on the characteristics of different datasets, optimizing the structure to align with the data. When the length of the
Softmax kernel $T_{\lambda}$ exceeds a certain value, ASRC-SNN shows reduced sensitivity to variations in $T_{\lambda}$, while its performance approaches the optimal level (see Section \ref{sec:Ablation_ASRC}). These characteristics enhance the robustness of ASRC-SNN, enabling it to preserve stability across diverse datasets and under fluctuations in hyperparameters. \par

\section{Experiments}
\label{Experiments}
\begin{table*}[ht]
\centering
\caption{Classification accuracy on S-MNIST, PS-MNIST, SSC and GSC datasets. The symbol * indicates that the results are derived from \cite{zhang2024tc-lif}.}
\footnotesize 
\setlength{\tabcolsep}{4pt} 
\renewcommand{\arraystretch}{0.9} 
\begin{tabular*}{\textwidth}{@{\extracolsep{\fill}}llccc@{\extracolsep{\fill}}}
\toprule
\textbf{Dataset} & \textbf{Method} & \textbf{Recurrent} & \textbf{Parameters} & \textbf{Accuracy(\%)} \\
\midrule
\multirow{8}{*}{\shortstack{S-MNIST \\ \footnotesize(timestep=784)}} & PLIF \cite{fang2021plif} & Y & 0.15M* & 91.79 \\
& GLIF \cite{yao2022glif} & Y & 0.15M* & 96.64 \\
& ALIF \cite{yin2021alif} & Y & 0.15M & 98.70 \\
& BRFN \cite{higuchi2024bhrf} & N & 0.068M & 99.1 \\
& TC-LIF \cite{zhang2024tc-lif} & Y & 0.063M/0.15M & 98.79/99.20 \\
& PMSN \cite{chen2024pmsn} & N & 0.066M/0.15M & 99.40/99.53 \\
& \textbf{SRC-SNN (ours)} & Y & \textbf{0.063M/0.15M} & \textbf{99.32/99.38} \\
& \textbf{ASRC-SNN (ours)} & Y & \textbf{0.063M} & \textbf{99.57} \\
\midrule
\multirow{7}{*}{\shortstack{PS-MNIST \\ \footnotesize(timestep=784)}} & GLIF \cite{yao2022glif} & Y & 0.15M* & 90.47 \\
& ALIF \cite{yin2021alif} & Y & 0.15M & 94.30 \\
& BRFN \cite{higuchi2024bhrf} & N & 0.068M & 95.2 \\
& TC-LIF \cite{zhang2024tc-lif} & Y & 0.063M/0.15M & 92.69/95.36 \\
& PMSN \cite{chen2024pmsn} & N & 0.066M/0.15M & 97.16/97.78 \\
& \textbf{SRC-SNN (ours)} & Y & \textbf{0.063M/0.15M} & \textbf{94.78/96.36} \\
& \textbf{ASRC-SNN (ours)} & Y & \textbf{0.063M/0.15M} & \textbf{95.40/96.62} \\
\midrule
\multirow{9}{*}{\shortstack{SSC \\ \footnotesize(timestep=250)}} & TC-LIF \cite{zhang2024tc-lif} & Y & 0.11M & 61.90 \\
& SNN-CNN \cite{sadovsky2023snn-cnn} & N & N/A & 72.03 \\
& ALIF \cite{yin2021alif} & Y & N/A & 74.20 \\
& SpikGRU \cite{dampfhoffer2022spikingGRU} & Y & 0.28M & 77.00 \\
& RadLIF \cite{bittar2022radlif} & Y & 3.9M & 77.40 \\
& DCLS-Delays (2L-1KC) \cite{hammouamri2024dcls} & N & 0.70M & 79.77 \\
& DCLS-Delays (3L-2KC) \cite{hammouamri2024dcls} & N & 2.5M & 80.69 \\
& \textbf{SRC-SNN (ours)} & Y & \textbf{0.37M} & \textbf{81.83} \\
& \textbf{ASRC-SNN (ours)} & Y & \textbf{0.37M} & \textbf{81.93} \\
\midrule
\multirow{5}{*}{\shortstack{GSC \\ \footnotesize(timestep=101)}} & SNN with SFA \cite{salaj2021SFA} & Y & 4.3M & 91.21 \\
& ALIF \cite{yin2021alif} & Y & 0.22M & 92.10 \\
& TC-LIF \cite{zhang2024tc-lif} & Y & 0.19M & 94.84 \\
& \textbf{SRC-SNN (ours)} & Y & \textbf{0.088M} & \textbf{96.18} \\
& \textbf{ASRC-SNN (ours)} & Y & \textbf{0.089M} & \textbf{96.29} \\
\bottomrule
\end{tabular*}
\label{tab:my_label}
\end{table*}

\subsection{Experimental Setup}
We chose to evaluate our method on various temporal classification benchmarks, including sequential MNIST (S-MNIST), permuted sequential MNIST (PS-MNIST), Google Speech Commands v0.01 (GSC) and Spiking Google Speech Commands (SSC). We use a simple model architecture consisting of three hidden layers. Furthermore, we evaluate the effectiveness of SRC and ASRC on the more challenging sequential CIFAR (SCIFAR) dataset using a deeper network with six hidden layers. Dataset description and more experimental details are provided in Appendix \ref{appd:training}. \par
 
\subsection{Resluts}
Table.\ref{tab:my_label} compares our two proposed methods, SRC-SNN and ASRC-SNN, with previous works in the field of SNNs on four benchmark datasets (S-MNIST, PS-MNIST, SSC and GSC) in terms of accuracy, model size and whether recurrent connections were used. SRC-SNN outperforms the previous state-of-the-art accuracy on the GSC and SSC benchmark datasets, while significantly reducing the number of parameters. ASRC-SNN further improves upon SRC-SNN, achieving state-of-the-art performance on S-MNIST, SSC and GSC. On the PS-MNIST dataset, ASRC-SNN surpasses other approaches that use recurrent structures.

\begin{figure*}[t]
    \centering
    
    \begin{subfigure}[t]{0.23\textwidth}
        \centering
        \includegraphics[width=1.0\textwidth]{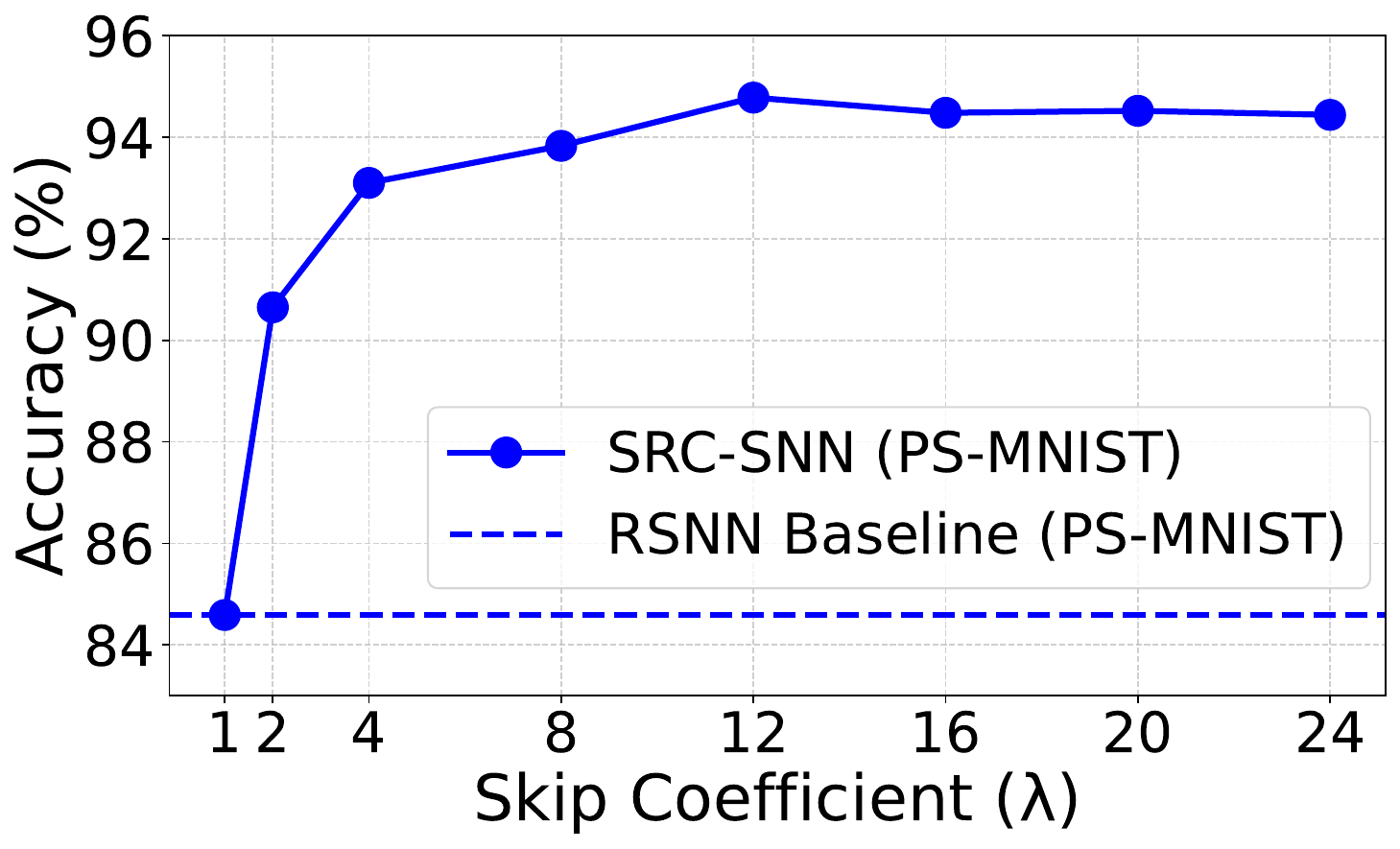}
        \centering
        \caption{PS-MNIST}
        \label{fig:lambda_psmnist}
    \end{subfigure}
    \hspace{2pt}
    \begin{subfigure}[t]{0.23\textwidth}
        \centering
        \includegraphics[width=1.0\textwidth]{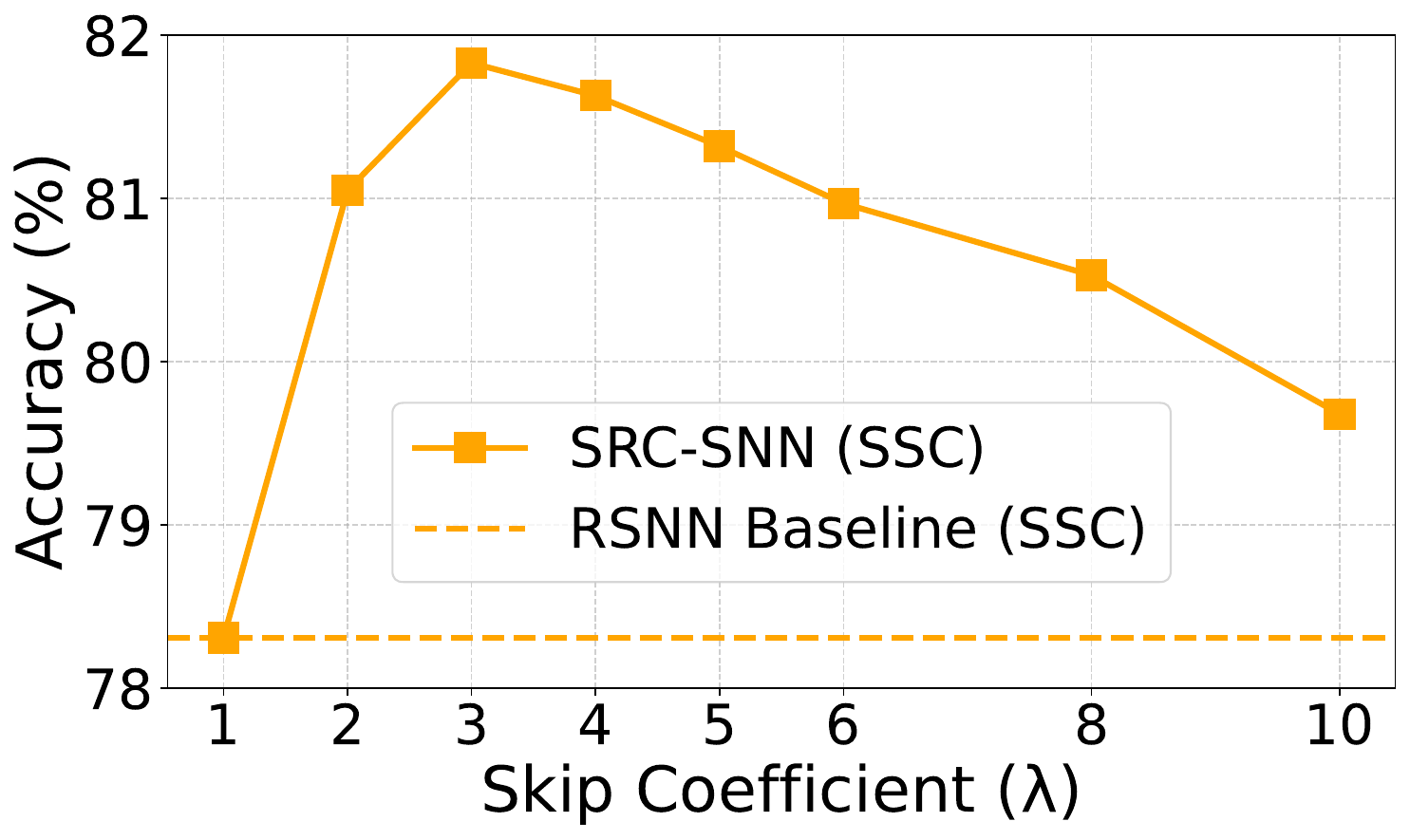}
        \caption{SSC}
        \label{fig:lambda_ssc}
        
    \end{subfigure}
    \hspace{2pt}
    \begin{subfigure}[t]{0.23\textwidth}
        \centering
        \includegraphics[width=1.0\textwidth]{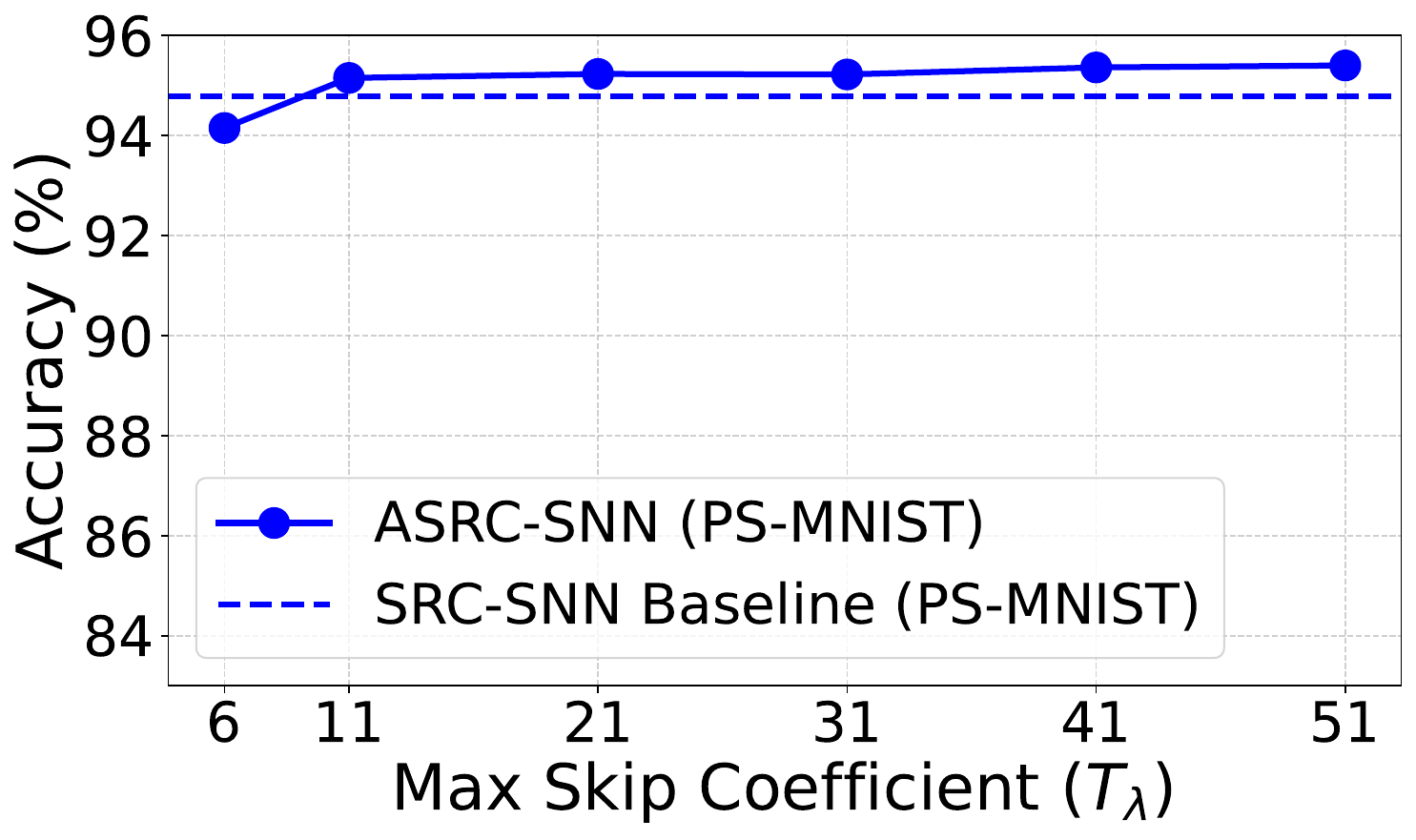}
        \caption{PS-MNIST}
        \label{fig:psmnist_performance}
    \end{subfigure}
    \hspace{2pt}
    \begin{subfigure}[t]{0.23\textwidth}
        \centering
        \includegraphics[width=1.0\textwidth]{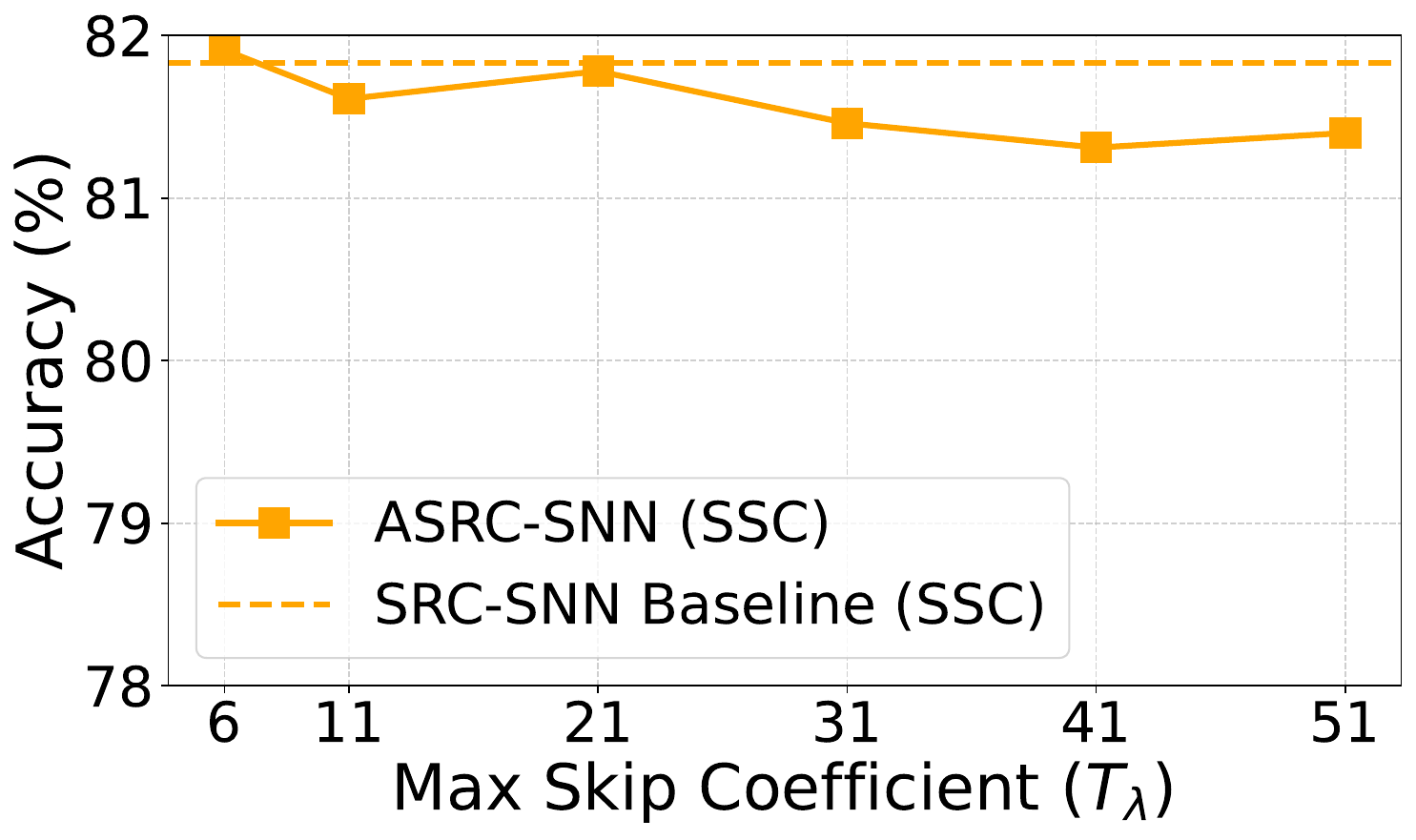}
        \caption{SSC}
        \label{fig:ssc_performance}
    \end{subfigure}
    
    \caption{
        Effects of $\lambda$/$T_{\lambda}$ on SRC-SNN/ASRC-SNN in SSC and PS-MNIST benchmarks. 
        (a)-(b): Impact of $\lambda$ on SRC-SNN; 
        (c)-(d): Impact of $T_{\lambda}$ on ASRC-SNN.
        All results are shown on PS-MNIST and SSC datasets respectively.
    }
    \label{fig:combined_analysis}
\end{figure*}

\subsection{Ablation and Analysis}
In this section, we conduct controlled experiments to investigate the effectiveness of the SRC and ASRC methods, analyzing the impact of the skip coefficient on SRC-SNN and the maximum skip coefficient on ASRC-SNN using relatively complex SSC and PS-MNIST datasets.
\subsubsection{skip coefficient on SRC-SNN}
\label{sec:Ablation_SRC}
The results presented in Figure.\ref{fig:lambda_psmnist} and Figure.\ref{fig:lambda_ssc} indicate that when the skip recurrent coefficient $\lambda$ of SRC-SNN exceeds 1, the model performance improves dramatically. This suggests that SRC-SNN demonstrates a significant improvement in long-term temporal modeling ability compared to the vanilla RSNN. However, it can be observed that the preferred values of $\lambda$ differ significantly between different datasets, with the preferred values for the SSC dataset being 3, 4 and 5, while those for PS-MNIST being 12, 16, 20 and 24. On the one hand, the optimal values $\lambda$ vary greatly between datasets; on the other hand, the sensitivity of the results to the setting of $\lambda$ also differs between datasets, with the SSC dataset being particularly sensitive to $\lambda$. These factors complicate the hyparameter search process in SRC-SNN. \par
\subsubsection{max skip coefficient on ASRC-SNN}
\label{sec:Ablation_ASRC}

To validate the robustness of ASRC-SNN, the range and variation of $T_{\lambda}$ in the experiments of this part are wider and more extensive compared to the values of $\lambda$ in Section \ref{sec:Ablation_SRC}. As observed in Figure.\ref{fig:psmnist_performance} and Figure.\ref{fig:ssc_performance}, once $T_{\lambda}$ exceeds a certain value, further increases in $T_{\lambda}$ result in slight performance fluctuations, with performance approaching the optimal level. This demonstrates the robustness of ASRC-SNN. Furthermore, the optimal performance of ASRC-SNN exceeds the best performance of the SRC-SNN model on both datasets. Notably, on the PS-MNIST dataset with longer sequences, the performance of ASRC-SNN consistently outperforms the best performance of SRC-SNN when $T_{\lambda}$ is set to 11 or higher. In summary, ASRC-SNN demonstrates superior temporal modeling capabilities and robustness compared to SRC-SNN. \par
\subsection{The Impact of Neurons on SRC-SNN/ASRC-SNN}
\begin{figure}[t] 
    \centering
    \begin{subfigure}[t]{0.23\textwidth}
        \centering
        \includegraphics[width=1.0\textwidth]{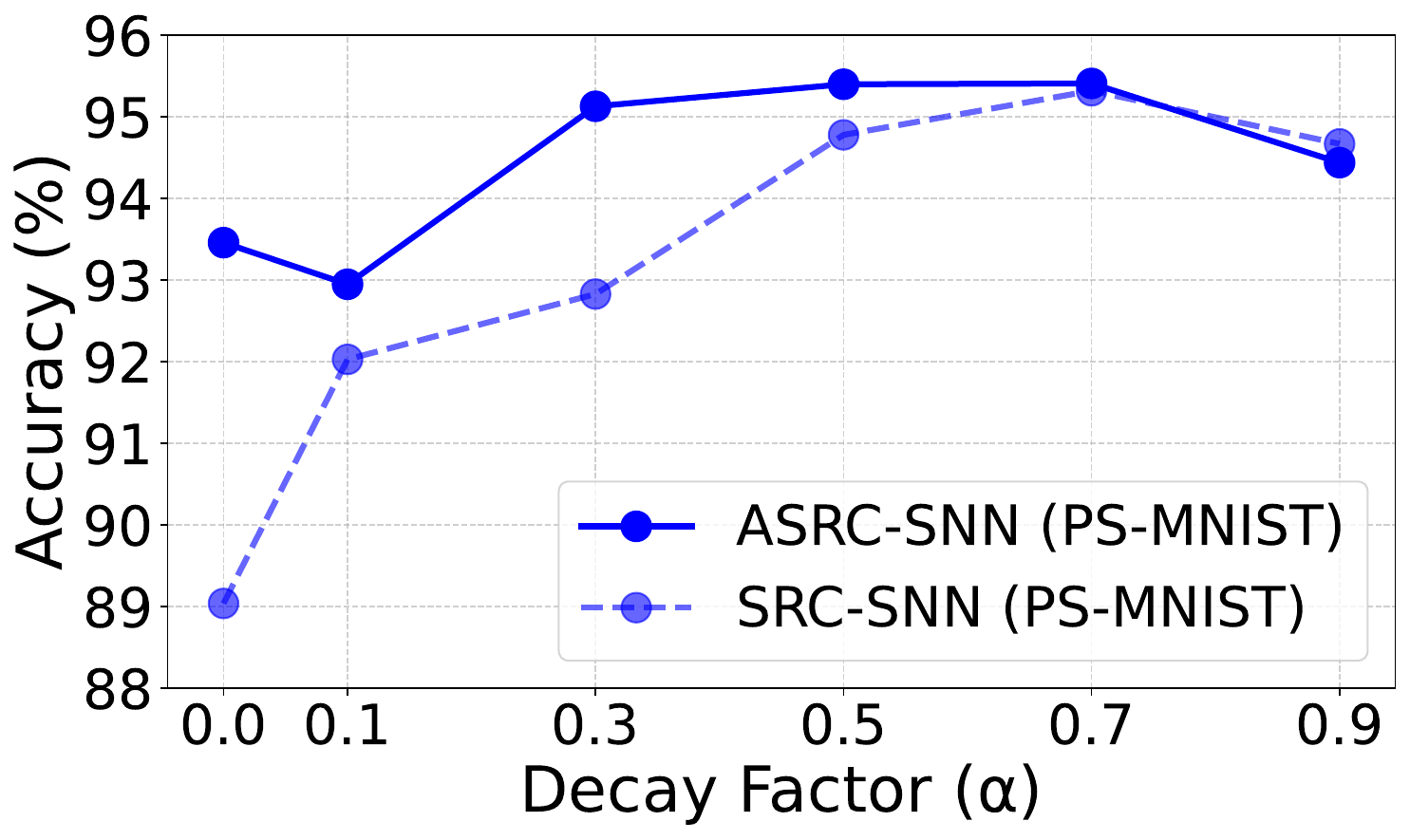}
         
        \caption{PS-MNIST}
        \label{fig:alpha_psmnist}
    \end{subfigure}
    \hspace{2pt}
    \begin{subfigure}[t]{0.23\textwidth}
        \centering
        \includegraphics[width=1.0\textwidth]{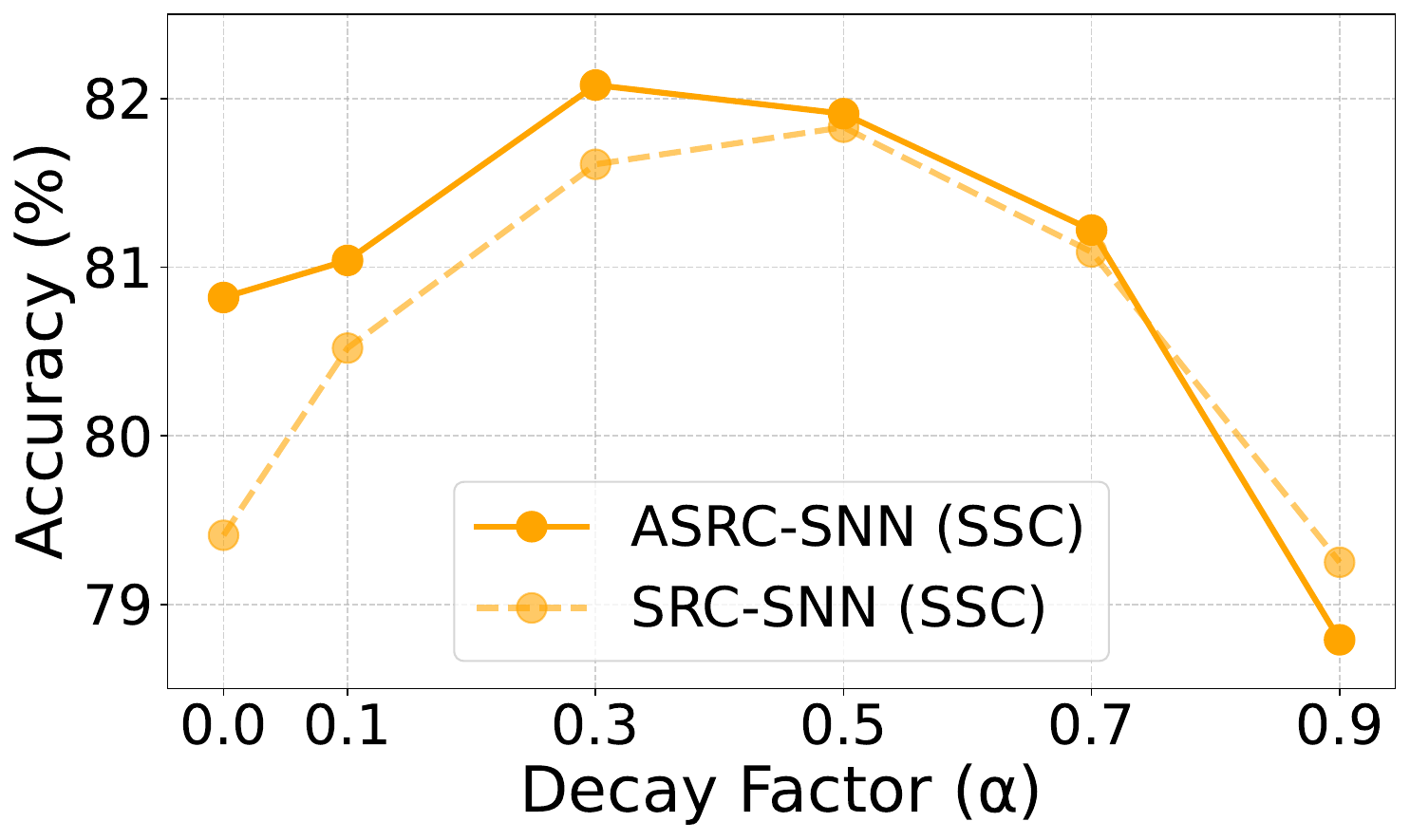}
        \caption{SSC}
        \label{fig:alpha_ssc}
    \end{subfigure}   
    \hspace{2pt}
    \begin{subfigure}[t]{0.23\textwidth}
        \centering
        \includegraphics[width=1.0\textwidth]{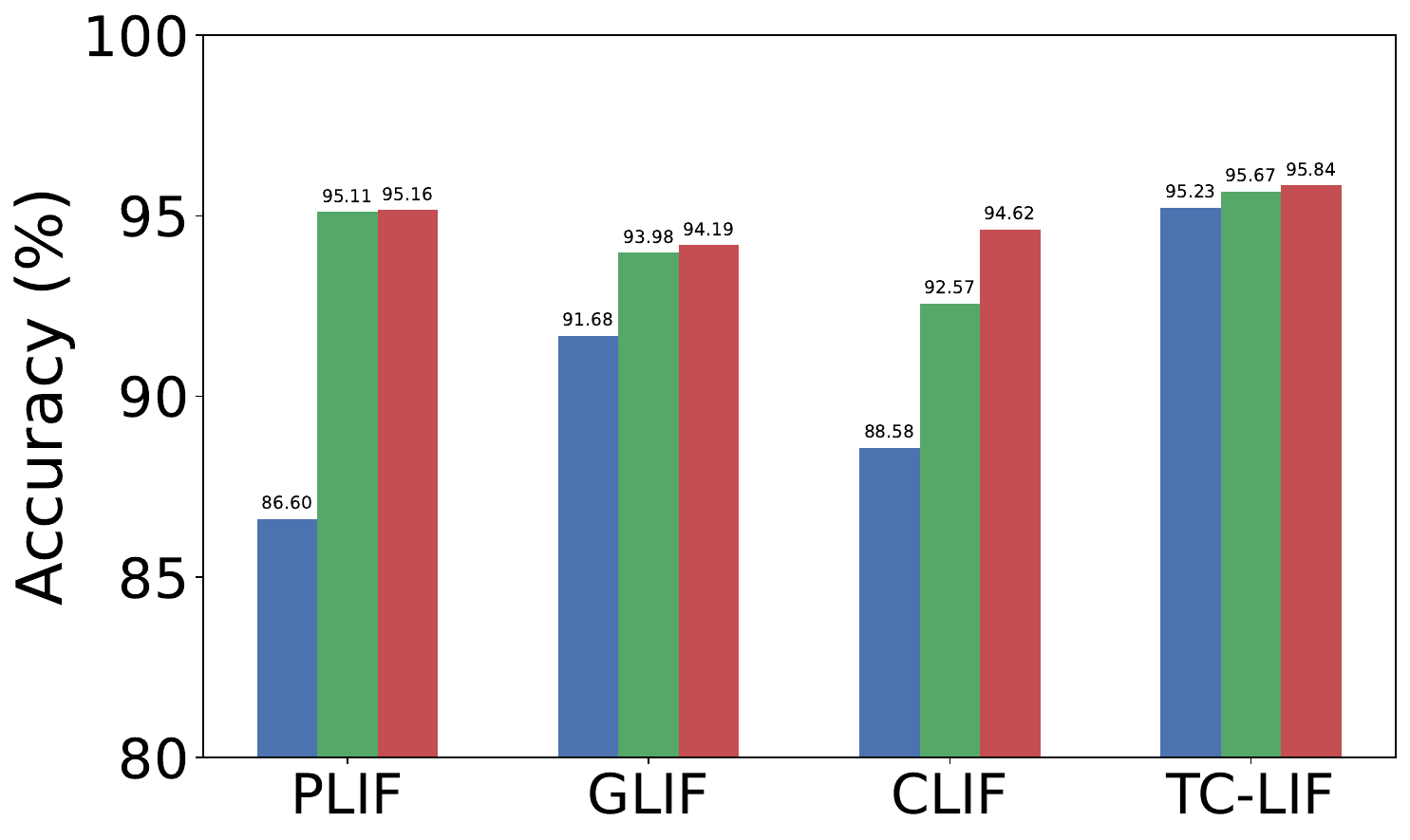}
        \caption{PS-MNIST}
        \label{fig:generality_psmnist}
    \end{subfigure}
    \hspace{2pt}
    \begin{subfigure}[t]{0.23\textwidth}
        \centering
        \includegraphics[width=1.0\textwidth]{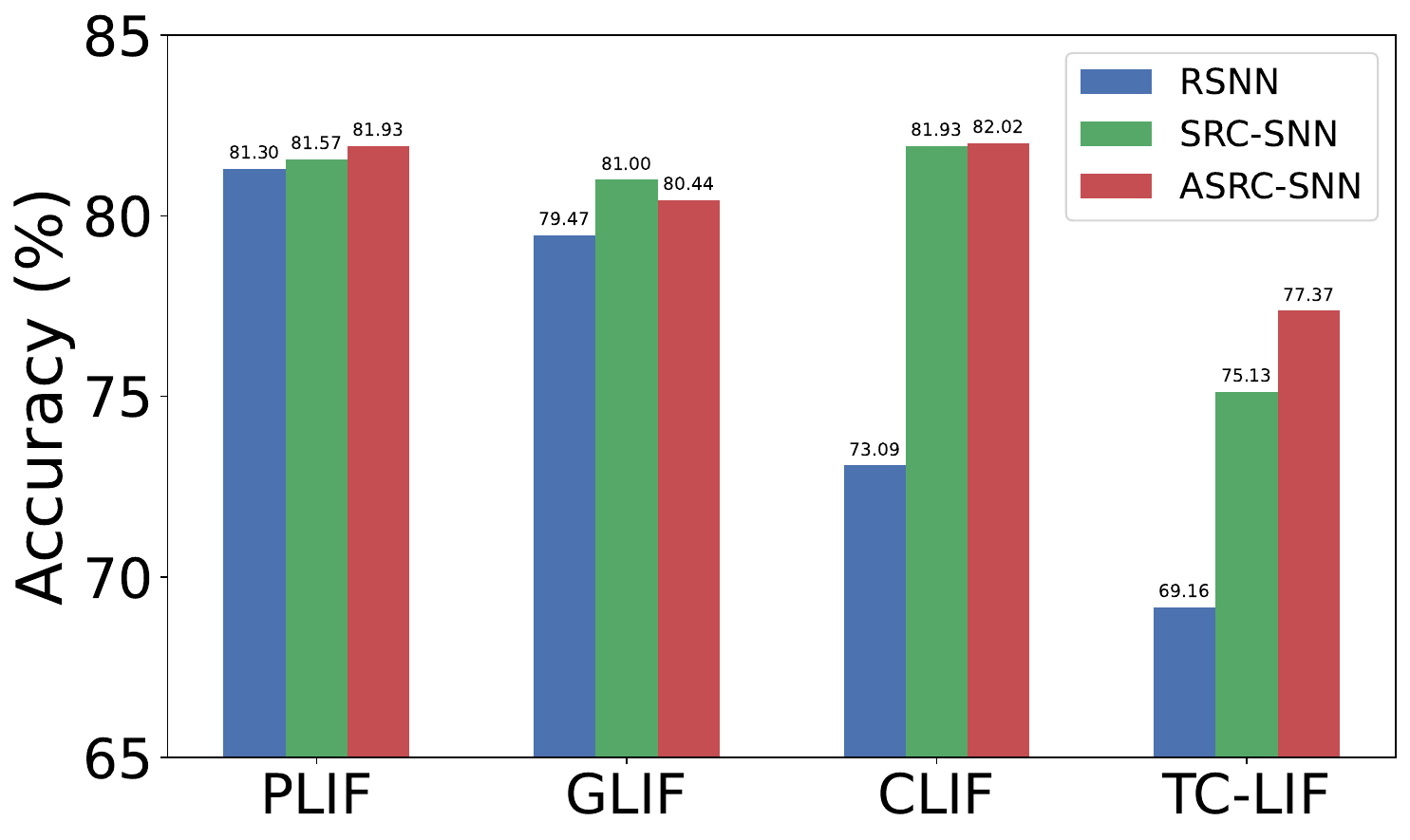}
        \caption{SSC}
        \label{fig:generality_ssc}
    \end{subfigure}   
    \caption{The impact of the membrane potential decay factor $\alpha$ in LIF neurons on the performance of ASRC-SNN across the PS-MNIST and SSC datasets. (a) and (b) correspond to the results on the PS-MNIST and SSC datasets, respectively.}
    \label{fig:lif}
\end{figure}

\subsubsection{the impact of LIF neuron}
As shown in Figure.\ref{fig:alpha_psmnist} and Figure.\ref{fig:alpha_ssc}, when the membrane potential decay factor $\alpha = 0$, the LIF neuron degenerates into the Heaviside function, leading to degraded performance compared to when $\alpha$ is within a reasonable range. This indicates that LIF neurons, in combination with skip recurrent connections, play a collaborative role in temporal modeling. Furthermore, when LIF neurons degenerate into the Heaviside function, the temporal modeling primarily relies on the SRC or ASRC modules. In this scenario, ASRC demonstrates a clear performance advantage over SRC. However, when neurons retain a certain degree of temporal expressiveness, the performance gap between SRC and ASRC narrows, highlighting the critical role of LIF neurons in the collaborative dynamics between the recurrent structure and neurons.

\subsubsection{generalize to other spiking neuron models}
In this section, we investigate the impact of replacing the LIF neuron with alternative neuron models, including PLIF\cite{fang2021plif}, GLIF\cite{yao2022glif}, CLIF\cite{huang2024clif} and TC-LIF\cite{zhang2024tc-lif}. We evaluate whether the SRC/ASRC architectures remain effective under these substitutions compared to vanilla RNN structures. As shown in Figures.\ref{fig:generality_psmnist} and Figures.\ref{fig:generality_ssc}, both SRC-SNN and ASRC-SNN consistently outperform standard RSNNs across all tested neuron types. Notably, PLIF and TC-LIF exhibit strong performance on one dataset (either PS-MNIST or SSC) while performing poorly on the other, indicating dataset-dependent behavior. However, when integrated into the SRC/ASRC framework, their performance on the weaker dataset improves significantly, demonstrating the robustness and adaptability of our architecture. For the CLIF neuron, which performs poorly on both datasets under a vanilla RSNN, our method brings substantial performance gains, even achieving the best result on SSC. Additionally, we observe that ASRC generally outperforms SRC across neuron types, with the only exception being the GLIF neuron on the SSC dataset.

\begin{figure}[t] 
    \centering
    \begin{subfigure}[t]{0.45\textwidth}
        \centering
        \includegraphics[width=1.0\textwidth]{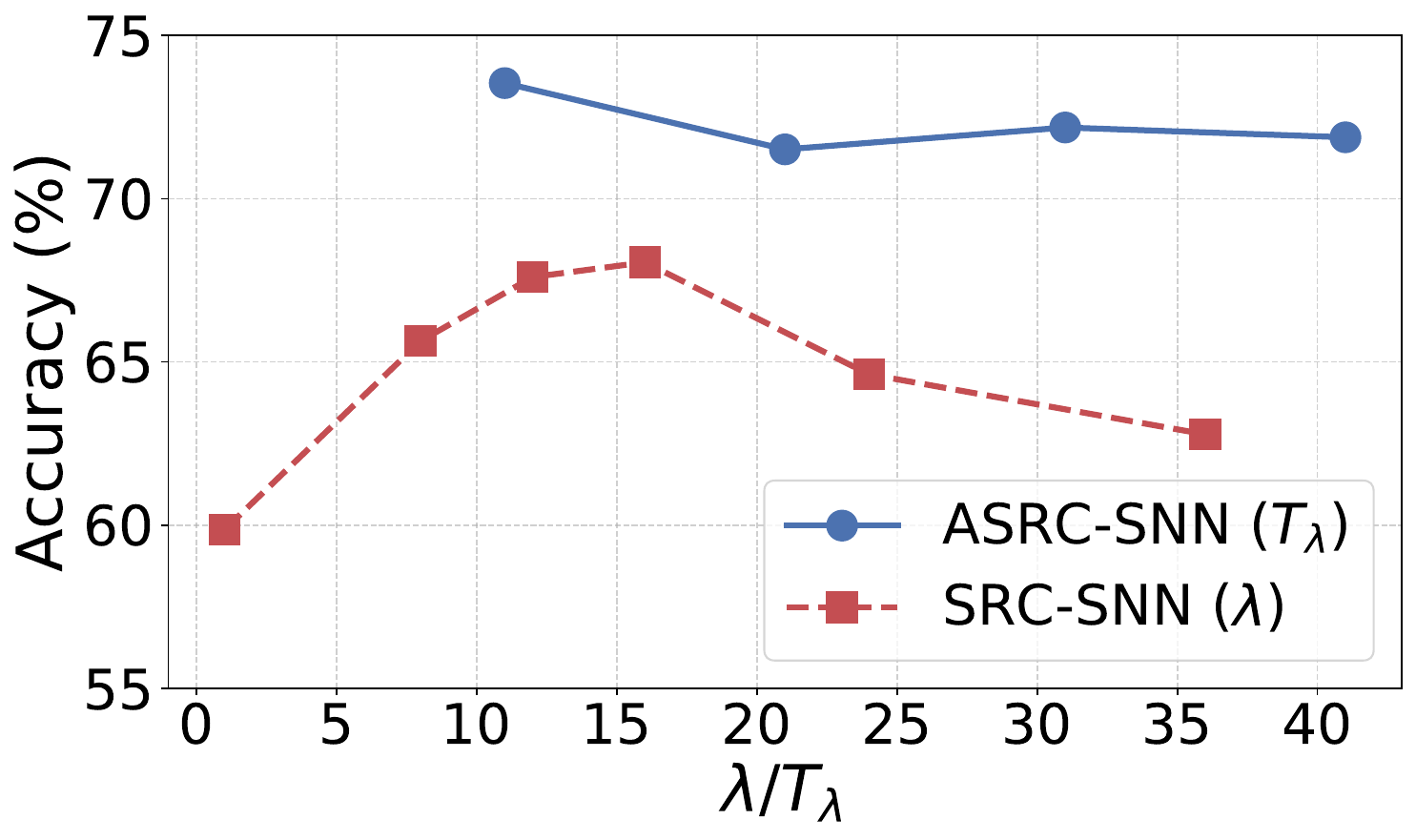}
        \caption{SCIFAR(timestep=1024)}
        \label{fig:scifar}
    \end{subfigure}
    \hspace{2pt}
    \begin{subfigure}[t]{0.45\textwidth}
        \centering
        \includegraphics[width=1.0\textwidth]{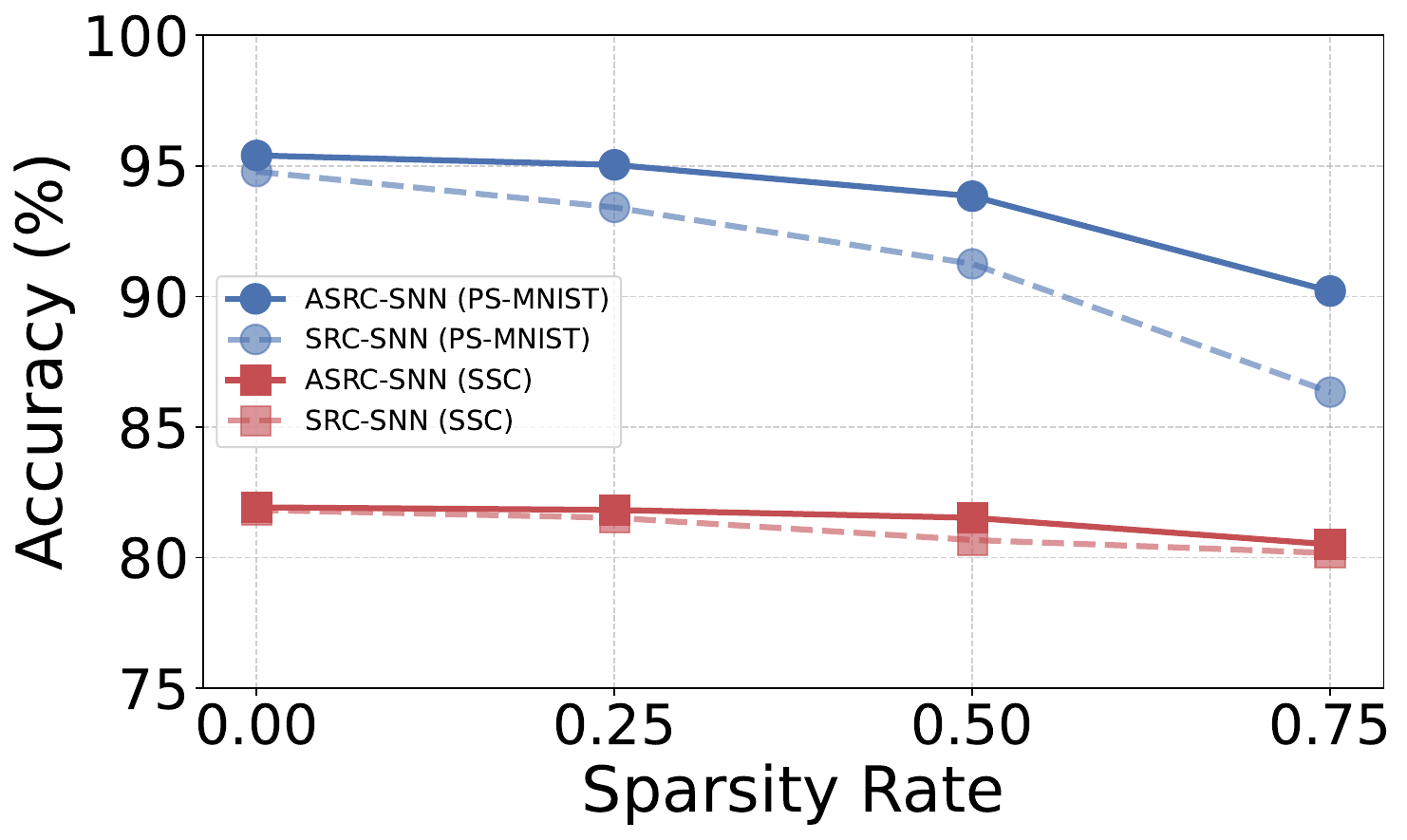}
        \caption{SRC-SNN/ASRC-SNN under diffirent sparsity}
        \label{fig:sparsity}
    \end{subfigure}    
    \caption{The comparison between SRC-SNN and ASRC-SNN under complex datasets and sparse connectivity conditions.}
    \label{fig:lif}
\end{figure}
\subsection{The superiority of ASRC over SRC under Challenging Conditions}
\textbf{Complex SCIFAR dataset} As shown in Figures.\ref{fig:scifar}, our SRC-SNN achieves more than an 8\% improvement in accuracy over the standard RSNN, while ASRC-SNN further improves performance by over 5\% compared to SRC-SNN, with enhanced robustness. This indicates that ASRC demonstrates more significant advantages over SRC on complex datasets.

\textbf{Sparse Connectivity} As shown in Figures.\ref{fig:sparsity}, across different sparsity levels, ASRC-SNN consistently outperforms SRC-SNN. This advantage is particularly evident on the PS-MNIST dataset, which has complex temporal dependencies—where increasing sparsity further amplifies ASRC-SNN’s superiority over SRC-SNN. Additionally, we observed that as sparsity increases, our model’s performance on SSC does not degrade significantly, which may be related to the tendency to overfit easily on this dataset.
\section{Discussion}
\label{Discussion}
In this work, we reveal the presence of a gradient vanishing problem along the temporal dimension by treating recurrent structures and spiking neurons within a unified framework. To address this issue, we propose Skip Recurrent Connections (SRC), and further introduce Adaptive SRC (ASRC), which demonstrates enhanced temporal modeling capabilities and robustness compared to SRC.

In Appendix.\ref{appd:Overhead}, we provide the computational, energy, and storage overhead analysis. Although ASRC introduces only a marginal increase in training time relative to RSNNs, the inherently high training cost of RSNNs makes it worthwhile to explore the parallelization of SRC and ASRC. When $\lambda$/$T_{\lambda}$ are large, the additional memory overhead during training becomes non-negligible. Nevertheless, we believe that our work represents a meaningful step forward in revisiting and extending SRC\cite{zhang2016skippedrnn}, especially in light of the strong robustness exhibited by ASRC.

The important role of neurons in the SRC-SNN/ASRC-SNN has driven the exploration of neuron models better suited for integration with SRC/ASRC. Moreover, our method bears intuitive resemblance to residual connections\cite{he2016residual}, suggesting that exploring adaptive skip connections in feedforward architectures could also be an interesting research direction. Finally, we have some thoughts for future research directions:

 1) The essence of ASRC lies in learning a discrete position along the temporal dimension, with the potential to extend this method to learning a discrete position in both time and space. To learn multiple positions, the ASRC method can be applied repeatedly. A promising application of this approach is the learning of non-zero positions in dilated convolution kernels \cite{yu2015dc}, similar to \cite{khalfaoui2021DCLS}.
 
2) We have observed that as the output of the Softmax function approaches Hardmax, the variance of the output becomes larger. Based on this observation, we plan to investigate a new approach for implementing ASRC: building upon the ASRC in this paper, by discarding the temperature parameter in the Softmax kernel function and incorporating the variance of the Softmax kernel output as part of the loss function.

3) Inspired by the sharp weight distribution of the Softmax kernel during the intermediate phase of ASRC-SNN training, we will explore the possibility of adaptive multi-skip recurrent connections, considering both parameter-shared and parameter-independent versions.

\nocite{*}
\bibliographystyle{unsrt}
\bibliography{example_paper}

\begin{thebibliography}{10}

\bibitem{roy2019sparse0}
Kaushik Roy, Akhilesh Jaiswal, and Priyadarshini Panda.
\newblock Towards spike-based machine intelligence with neuromorphic computing.
\newblock {\em Nature}, 575(7784):607--617, 2019.

\bibitem{nunes2022sparse}
Joao~D Nunes, Marcelo Carvalho, Diogo Carneiro, and Jaime~S Cardoso.
\newblock Spiking neural networks: A survey.
\newblock {\em IEEE Access}, 10:60738--60764, 2022.

\bibitem{akopyan2015chip}
Filipp Akopyan, Jun Sawada, Andrew Cassidy, Rodrigo Alvarez-Icaza, John Arthur, Paul Merolla, Nabil Imam, Yutaka Nakamura, Pallab Datta, Gi-Joon Nam, et~al.
\newblock Truenorth: Design and tool flow of a 65 mw 1 million neuron programmable neurosynaptic chip.
\newblock {\em IEEE transactions on computer-aided design of integrated circuits and systems}, 34(10):1537--1557, 2015.

\bibitem{davies2018chip}
Mike Davies, Narayan Srinivasa, Tsung-Han Lin, Gautham Chinya, Yongqiang Cao, Sri~Harsha Choday, Georgios Dimou, Prasad Joshi, Nabil Imam, Shweta Jain, et~al.
\newblock Loihi: A neuromorphic manycore processor with on-chip learning.
\newblock {\em Ieee Micro}, 38(1):82--99, 2018.

\bibitem{pei2019chip}
Jing Pei, Lei Deng, Sen Song, Mingguo Zhao, Youhui Zhang, Shuang Wu, Guanrui Wang, Zhe Zou, Zhenzhi Wu, Wei He, et~al.
\newblock Towards artificial general intelligence with hybrid tianjic chip architecture.
\newblock {\em Nature}, 572(7767):106--111, 2019.

\bibitem{gerstner2002book_lif}
Wulfram Gerstner and Werner~M Kistler.
\newblock {\em Spiking neuron models: Single neurons, populations, plasticity}.
\newblock Cambridge university press, 2002.

\bibitem{ding2021conversion}
Jianhao Ding, Zhaofei Yu, Yonghong Tian, and Tiejun Huang.
\newblock Optimal ann-snn conversion for fast and accurate inference in deep spiking neural networks.
\newblock {\em arXiv preprint arXiv:2105.11654}, 2021.

\bibitem{zhou2023spikformer}
Zhaokun Zhou, Yuesheng Zhu, Chao He, Yaowei Wang, Shuicheng YAN, Yonghong Tian, and Li~Yuan.
\newblock Spikformer: When spiking neural network meets transformer.
\newblock In {\em The Eleventh International Conference on Learning Representations}, 2023.

\bibitem{yao2024sdriven}
Man Yao, Jiakui Hu, Zhaokun Zhou, Li~Yuan, Yonghong Tian, Bo~Xu, and Guoqi Li.
\newblock Spike-driven transformer.
\newblock {\em Advances in neural information processing systems}, 36, 2024.

\bibitem{zhou2024qkformer}
Chenlin Zhou, Han Zhang, Zhaokun Zhou, Liutao Yu, Liwei Huang, Xiaopeng Fan, Li~Yuan, Zhengyu Ma, Huihui Zhou, and Yonghong Tian.
\newblock {QKF}ormer: Hierarchical spiking transformer using q-k attention.
\newblock In {\em The Thirty-eighth Annual Conference on Neural Information Processing Systems}, 2024.

\bibitem{li2017cifar10}
Hongmin Li, Hanchao Liu, Xiangyang Ji, Guoqi Li, and Luping Shi.
\newblock Cifar10-dvs: an event-stream dataset for object classification.
\newblock {\em Frontiers in neuroscience}, 11:309, 2017.

\bibitem{amir2017dvs128}
Arnon Amir, Brian Taba, David Berg, Timothy Melano, Jeffrey McKinstry, Carmelo Di~Nolfo, Tapan Nayak, Alexander Andreopoulos, Guillaume Garreau, Marcela Mendoza, et~al.
\newblock A low power, fully event-based gesture recognition system.
\newblock In {\em Proceedings of the IEEE conference on computer vision and pattern recognition}, pages 7243--7252, 2017.

\bibitem{lenero2011dvs}
Juan~Antonio Le{\~n}ero-Bardallo, Teresa Serrano-Gotarredona, and Bernab{\'e} Linares-Barranco.
\newblock A 3.6 $\mu$s latency asynchronous frame-free event-driven dynamic-vision-sensor.
\newblock {\em IEEE Journal of Solid-State Circuits}, 46(6):1443--1455, 2011.

\bibitem{deng2023surrogate}
Shikuang Deng, Hao Lin, Yuhang Li, and Shi Gu.
\newblock Surrogate module learning: Reduce the gradient error accumulation in training spiking neural networks.
\newblock In {\em International Conference on Machine Learning}, pages 7645--7657. PMLR, 2023.

\bibitem{ma2023NSNN}
Gehua Ma, Rui Yan, and Huajin Tang.
\newblock Exploiting noise as a resource for computation and learning in spiking neural networks.
\newblock {\em Patterns}, 2023.

\bibitem{wang2023ASGL}
Ziming Wang, Runhao Jiang, Shuang Lian, Rui Yan, and Huajin Tang.
\newblock Adaptive smoothing gradient learning for spiking neural networks.
\newblock In {\em International Conference on Machine Learning}, pages 35798--35816. PMLR, 2023.

\bibitem{huang2024clif}
Yulong Huang, Xiaopeng Lin, Hongwei Ren, Haotian Fu, Yue Zhou, Zunchang Liu, Biao Pan, and Bojun Cheng.
\newblock Clif: Complementary leaky integrate-and-fire neuron for spiking neural networks.
\newblock {\em arXiv preprint arXiv:2402.04663}, 2024.

\bibitem{yin2021alif}
Bojian Yin, Federico Corradi, and Sander~M Boht{\'e}.
\newblock Accurate and efficient time-domain classification with adaptive spiking recurrent neural networks.
\newblock {\em Nature Machine Intelligence}, 3(10):905--913, 2021.

\bibitem{elman1990vanillarnn}
Jeffrey~L Elman.
\newblock Finding structure in time.
\newblock {\em Cognitive science}, 14(2):179--211, 1990.

\bibitem{bittar2022radlif}
Alexandre Bittar and Philip~N Garner.
\newblock A surrogate gradient spiking baseline for speech command recognition.
\newblock {\em Frontiers in Neuroscience}, 16:865897, 2022.

\bibitem{zhang2024tc-lif}
Shimin Zhang, Qu~Yang, Chenxiang Ma, Jibin Wu, Haizhou Li, and Kay~Chen Tan.
\newblock Tc-lif: A two-compartment spiking neuron model for long-term sequential modelling.
\newblock In {\em Proceedings of the AAAI Conference on Artificial Intelligence}, volume~38, pages 16838--16847, 2024.

\bibitem{baronig2024advancing}
Maximilian Baronig, Romain Ferrand, Silvester Sabathiel, and Robert Legenstein.
\newblock Advancing spatio-temporal processing in spiking neural networks through adaptation.
\newblock {\em arXiv preprint arXiv:2408.07517}, 2024.

\bibitem{henaff2016orthogonal}
Mikael Henaff, Arthur Szlam, and Yann LeCun.
\newblock Recurrent orthogonal networks and long-memory tasks.
\newblock In {\em International Conference on Machine Learning}, pages 2034--2042. PMLR, 2016.

\bibitem{dampfhoffer2022spikingGRU}
Manon Dampfhoffer, Thomas Mesquida, Alexandre Valentian, and Lorena Anghel.
\newblock Investigating current-based and gating approaches for accurate and energy-efficient spiking recurrent neural networks.
\newblock In {\em International Conference on Artificial Neural Networks}, pages 359--370. Springer, 2022.

\bibitem{wang2024STC}
Lihao Wang and Zhaofei Yu.
\newblock Autaptic synaptic circuit enhances spatio-temporal predictive learning of spiking neural networks.
\newblock {\em arXiv preprint arXiv:2406.00405}, 2024.

\bibitem{fang2024PSN}
Wei Fang, Zhaofei Yu, Zhaokun Zhou, Ding Chen, Yanqi Chen, Zhengyu Ma, Timoth{\'e}e Masquelier, and Yonghong Tian.
\newblock Parallel spiking neurons with high efficiency and ability to learn long-term dependencies.
\newblock {\em Advances in Neural Information Processing Systems}, 36, 2024.

\bibitem{chen2024pmsn}
Xinyi Chen, Jibin Wu, Chenxiang Ma, Yinsong Yan, Yujie Wu, and Kay~Chen Tan.
\newblock Pmsn: A parallel multi-compartment spiking neuron for multi-scale temporal processing.
\newblock {\em arXiv preprint arXiv:2408.14917}, 2024.

\bibitem{higuchi2024bhrf}
Saya Higuchi, Sebastian Kairat, Sander~M Boht{\'e}, and Sebastian Otte.
\newblock Balanced resonate-and-fire neurons.
\newblock {\em arXiv preprint arXiv:2402.14603}, 2024.

\bibitem{hammouamri2024dcls}
Ilyass Hammouamri, Ismail Khalfaoui-Hassani, and Timoth{\'e}e Masquelier.
\newblock Learning delays in spiking neural networks using dilated convolutions with learnable spacings.
\newblock In {\em The Twelfth International Conference on Learning Representations}, 2024.

\bibitem{yao2021TA-SNN}
Man Yao, Huanhuan Gao, Guangshe Zhao, Dingheng Wang, Yihan Lin, Zhaoxu Yang, and Guoqi Li.
\newblock Temporal-wise attention spiking neural networks for event streams classification.
\newblock In {\em Proceedings of the IEEE/CVF International Conference on Computer Vision}, pages 10221--10230, 2021.

\bibitem{sadovsky2023snn-cnn}
Erik Sadovsky, Maros Jakubec, and Roman Jarina.
\newblock Speech command recognition based on convolutional spiking neural networks.
\newblock In {\em 2023 33rd International Conference Radioelektronika (RADIOELEKTRONIKA)}, pages 1--5. IEEE, 2023.

\bibitem{liu2024lmuformer}
Zeyu Liu, Gourav Datta, Anni Li, and Peter~Anthony Beerel.
\newblock Lmuformer: Low complexity yet powerful spiking model with legendre memory units.
\newblock {\em arXiv preprint arXiv:2402.04882}, 2024.

\bibitem{stan2024spikeSSM0}
Matei-Ioan Stan and Oliver Rhodes.
\newblock Learning long sequences in spiking neural networks.
\newblock {\em Scientific Reports}, 14(1):21957, 2024.

\bibitem{shen2024spikingssms}
Shuaijie Shen, Chao Wang, Renzhuo Huang, Yan Zhong, Qinghai Guo, Zhichao Lu, Jianguo Zhang, and Luziwei Leng.
\newblock Spikingssms: Learning long sequences with sparse and parallel spiking state space models.
\newblock {\em arXiv preprint arXiv:2408.14909}, 2024.

\bibitem{rueckauer2017softreset}
Bodo Rueckauer, Iulia-Alexandra Lungu, Yuhuang Hu, Michael Pfeiffer, and Shih-Chii Liu.
\newblock Conversion of continuous-valued deep networks to efficient event-driven networks for image classification.
\newblock {\em Frontiers in neuroscience}, 11:682, 2017.

\bibitem{han2020ssoftreset}
Bing Han, Gopalakrishnan Srinivasan, and Kaushik Roy.
\newblock Rmp-snn: Residual membrane potential neuron for enabling deeper high-accuracy and low-latency spiking neural network.
\newblock In {\em Proceedings of the IEEE/CVF conference on computer vision and pattern recognition}, pages 13558--13567, 2020.

\bibitem{neftci2019surrogate}
Emre~O Neftci, Hesham Mostafa, and Friedemann Zenke.
\newblock Surrogate gradient learning in spiking neural networks: Bringing the power of gradient-based optimization to spiking neural networks.
\newblock {\em IEEE Signal Processing Magazine}, 36(6):51--63, 2019.

\bibitem{deng2022TET}
Shikuang Deng, Yuhang Li, Shanghang Zhang, and Shi Gu.
\newblock Temporal efficient training of spiking neural network via gradient re-weighting.
\newblock {\em arXiv preprint arXiv:2202.11946}, 2022.

\bibitem{pascanu2013gradclip}
R~Pascanu.
\newblock On the difficulty of training recurrent neural networks.
\newblock {\em arXiv preprint arXiv:1211.5063}, 2013.

\bibitem{zhang2016skippedrnn}
Saizheng Zhang, Yuhuai Wu, Tong Che, Zhouhan Lin, Roland Memisevic, Russ~R Salakhutdinov, and Yoshua Bengio.
\newblock Architectural complexity measures of recurrent neural networks.
\newblock {\em Advances in neural information processing systems}, 29, 2016.

\bibitem{guo2017calibration}
Chuan Guo, Geoff Pleiss, Yu~Sun, and Kilian~Q Weinberger.
\newblock On calibration of modern neural networks.
\newblock In {\em International conference on machine learning}, pages 1321--1330. PMLR, 2017.

\bibitem{fang2021plif}
Wei Fang, Zhaofei Yu, Yanqi Chen, Timoth{\'e}e Masquelier, Tiejun Huang, and Yonghong Tian.
\newblock Incorporating learnable membrane time constant to enhance learning of spiking neural networks.
\newblock In {\em Proceedings of the IEEE/CVF international conference on computer vision}, pages 2661--2671, 2021.

\bibitem{yao2022glif}
Xingting Yao, Fanrong Li, Zitao Mo, and Jian Cheng.
\newblock Glif: A unified gated leaky integrate-and-fire neuron for spiking neural networks.
\newblock {\em Advances in Neural Information Processing Systems}, 35:32160--32171, 2022.

\bibitem{salaj2021SFA}
Darjan Salaj, Anand Subramoney, Ceca Kraisnikovic, Guillaume Bellec, Robert Legenstein, and Wolfgang Maass.
\newblock Spike frequency adaptation supports network computations on temporally dispersed information.
\newblock {\em Elife}, 10:e65459, 2021.

\bibitem{he2016residual}
Kaiming He, Xiangyu Zhang, Shaoqing Ren, and Jian Sun.
\newblock Deep residual learning for image recognition.
\newblock In {\em Proceedings of the IEEE conference on computer vision and pattern recognition}, pages 770--778, 2016.

\bibitem{yu2015dc}
F~Yu.
\newblock Multi-scale context aggregation by dilated convolutions.
\newblock {\em arXiv preprint arXiv:1511.07122}, 2015.

\bibitem{khalfaoui2021DCLS}
Ismail Khalfaoui-Hassani, Thomas Pellegrini, and Timoth{\'e}e Masquelier.
\newblock Dilated convolution with learnable spacings.
\newblock {\em arXiv preprint arXiv:2112.03740}, 2021.

\bibitem{maass1997lif}
Wolfgang Maass.
\newblock Networks of spiking neurons: the third generation of neural network models.
\newblock {\em Neural networks}, 10(9):1659--1671, 1997.

\bibitem{khalfaoui-hassani2023dilated}
Ismail Khalfaoui-Hassani, Thomas Pellegrini, and Timoth{\'e}e Masquelier.
\newblock Dilated convolution with learnable spacings: beyond bilinear interpolation.
\newblock In {\em ICML 2023 Workshop on Differentiable Almost Everything: Differentiable Relaxations, Algorithms, Operators, and Simulators}, 2023.

\bibitem{warden2018gsc}
Pete Warden.
\newblock Speech commands: A dataset for limited-vocabulary speech recognition.
\newblock {\em arXiv preprint arXiv:1804.03209}, 2018.

\bibitem{cramer2020ssc}
Benjamin Cramer, Yannik Stradmann, Johannes Schemmel, and Friedemann Zenke.
\newblock The heidelberg spiking data sets for the systematic evaluation of spiking neural networks.
\newblock {\em IEEE Transactions on Neural Networks and Learning Systems}, 33(7):2744--2757, 2020.

\bibitem{chung2014GRu}
Junyoung Chung, Caglar Gulcehre, KyungHyun Cho, and Yoshua Bengio.
\newblock Empirical evaluation of gated recurrent neural networks on sequence modeling.
\newblock {\em arXiv preprint arXiv:1412.3555}, 2014.

\bibitem{huang2024prf}
Yulong Huang, Zunchang Liu, Changchun Feng, Xiaopeng Lin, Hongwei Ren, Haotian Fu, Yue Zhou, Hong Xing, and Bojun Cheng.
\newblock Prf: Parallel resonate and fire neuron for long sequence learning in spiking neural networks.
\newblock {\em arXiv preprint arXiv:2410.03530}, 2024.

\end{thebibliography}


\newpage
\appendix
\onecolumn
\section{Dataset Description and Training Configuration}
\label{appd:training}
\subsection{Dataset Description}
\textit{\textbf{S-MNIST, PS-MNIST}} The MNIST dataset consists of 70,000 handwritten grayscale digit images with a resolution of 28×28 pixels, intended for classification tasks. Of these, 60,000 images are used for training, while 10,000 images are used for testing. In the S-MNIST dataset, the MNIST images are transformed into 784×1 vectors, where 784 represents the length of the temporal dimension. Building upon S-MNIST, the PS-MNIST dataset introduces random shuffling of the image sequences before inputting them into the network model, thereby creating more complex temporal dependencies compared to S-MNIST. \par
\textit{\textbf{SSC}} The SSC is a spike-based speech classification benchmark derived from Google Speech Commands v0.02, which contains 35 classes, proposed in \cite{cramer2020ssc}. The original waveform data have been converted into spike trains across 700 input channels. The dataset is divided into training, validation, and test splits, consisting of 75,466, 9,981 and 20,382 examples, respectively. The data were further processed with a discrete time scale of 5.6 ms to obtain a sequence length of 250 with zero right-padding. Additionally, the number of input neurons was reduced from 700 to 140 by binning every 5 neurons.\par
\textit{\textbf{GSC}} The GSC dataset consists of 64,727 audio files, which are divided into training, validation and test sets, containing 51,093, 6,799 and 3,081 samples, respectively, proposed
in \cite{warden2018gsc}. Our data preprocessing approach follows the TC-LIF procedure \cite{zhang2024tc-lif} , wherein the audio signals are first transformed into Mel-spectrograms and then converted to decibel units (dB).

\textit{\textbf{SCIFAR}} CIFAR-10 is a well-established image classification dataset consisting of 10 categories, with 6,000 color images per category. All images are of size 32×32 pixels, resulting in a total of 60,000 images—50,000 for training and 10,000 for testing. In the Sequential CIFAR (SCIFAR) dataset, the original images are flattened into time sequences of length 1024, where each time step contains the RGB pixel values. This transformation results in a 1024×3 sequential input, effectively converting the spatial image data into a temporal sequence. Such a representation increases the difficulty of the task and is particularly suitable for evaluating neural networks designed for sequential processing.

\begin{table*}[ht]
\centering
\caption{Hyperparameters used in different tasks.}
\footnotesize 
\setlength{\tabcolsep}{4pt} 
\renewcommand{\arraystretch}{0.9} 

\begin{tabular*}{\textwidth}{@{\extracolsep{\fill}}ccccccccc@{\extracolsep{\fill}}}
\toprule
\textbf{Dataset} & \textbf{Learning Rate} & \textbf{Softmax kernel learning rate} & \textbf{Weight Decay} & \textbf{Dropout} & \textbf{Batchsize} & \textbf{Epoch} \\
\midrule
S-MNIST & 0.001 & 0.1 & 0.01 & 0 & 256 & 200  \\
PS-MNIST & 0.001 & 0.1 & 0.01 & 0 & 256 & 200  \\
SSC & 0.001 & 0.1 & 0 & 0.1 & 128 & 100  \\
GSC & 0.0025 & 0.25 & 0 & 0.1 & 128 & 100  \\
SCIFAR & 0.0005 & 0.05 & 0.01 & 0 & 128 & 200  \\
\bottomrule
\end{tabular*}
\label{tab:conf1}
\end{table*}

\begin{table*}[ht]
\centering
\caption{Hyperparameters used in best models.}
\footnotesize 
\setlength{\tabcolsep}{4pt} 
\renewcommand{\arraystretch}{0.9} 
\begin{tabular*}{\textwidth}{@{\extracolsep{\fill}}cccccc@{\extracolsep{\fill}}}
\toprule
\textbf{Dataset} & \textbf{Model} & \textbf{Parameter} & \textbf{Hidden Size} & \textbf{$\lambda$} & \textbf{$T_{\lambda}$}\\
\midrule
\multirow{3}{*}{S-MNIST} & SRC-SNN & 0.063M & [64, 128, 128] & 16 & --   \\
& SRC-SNN & 0.15M & [64, 212, 212] & 12 & -- \\
& ASRC-SNN & 0.063M & [64, 128, 128] & - & 41 \\
\midrule
\multirow{4}{*}{PS-MNIST}  & SRC-SNN & 0.063M & [64, 128, 128] & 12 & --   \\
& SRC-SNN & 0.15M & [64, 212, 212] & 16 & --   \\
& ASRC-SNN & 0.063M & [64, 128, 128] & -- & 51   \\
& ASRC-SNN & 0.15M & [64, 212, 212] & -- & 31   \\
\midrule
\multirow{2}{*}{SSC} & SRC-SNN & 0.37M & [256, 256, 256] & 3 & --  \\
& ASRC-SNN & 0.37M & [256, 256, 256] & - & 6 \\
\midrule
\multirow{2}{*}{GSC} & SRC-SNN & 0.088M & [128, 128, 128] & 4 & --  \\
& ASRC-SNN & 0.089M & [128, 128, 128] & -- & 21 \\
\midrule
\multirow{2}{*}{SCIFAR} & SRC-SNN & 0.18M & [128, 128, 128, 128, 128, 128] & 16 & --  \\
& ASRC-SNN & 0.18M & [128, 128, 128, 128, 128, 128] & -- & 11 \\
\bottomrule
\end{tabular*}
\label{tab:conf2}
\end{table*}

\subsection{Training Configuration}
In our experiments, we use simple fully connected networks combined with the recurrent structure, without residual connections. Only on the SCIFAR dataset does our architecture use BatchNorm; no normalization is applied elsewhere. For the PS-MNIST, S-MNIST and SCFAR datasets, we employed the AdamW optimizer, while for the SSC and GSC datasets, we use the Adam optimizer. For SCIFAR, we use the CosineAnnealing learning rate scheduler; for other datasets, we apply the OneCycle learning rate schedule. For the softmax kernel parameters, we did not use weight decay, and the learning rate for the softmax kernel is set to 100× the global learning rate. Max learning rate and more detailed hyperparameter configurations can be found in Table.\ref{tab:conf1}. In addition, the hyperparameter configurations corresponding to the best performance of our models on different datasets are provided in Table.\ref{tab:conf2}.

\section{Computational, Energy, and Storage Overhead Analysis}
\label{appd:Overhead}
\textit{\textbf{Training Overhead}} As shown in Table.\ref{tab:training overhead}, the increase in memory consumption with larger values of $\lambda$/$T_{\lambda}$ is significant, while the increase in training time is relatively marginal. The training time of ASRC-SNN is approximately 17\% longer than SRC-SNN. When $\lambda$
 and $T_{\lambda}$ are close, the memory consumption of ASRC-SNN is slightly higher than SRC-SNN. Considering the trade-off between computational overhead and performance, we recommend selecting relatively small values of $\lambda$/$T_{\lambda}$
 whenever possible. In this case, selecting $\lambda = 8$
 and $T_{\lambda} = 11$
 results in good model performance, while the additional computational overhead remains acceptable compared to the vanilla RSNN. \par
\textit{\textbf{Energy and Storage Analysis}} Compared to vanilla RSNNs, SRC-SNN and ASRC-SNN achieve performance improvements solely by modifying the span of recurrent connections, without introducing additional computational complexity. As a result, their energy consumption remains the same. When considering deployment on neuromorphic hardware, from the perspective of individual neurons, SRC-SNN and ASRC-SNN only need to extra store spike information corresponding to a limited number of connection spans. Since each spike requires only 1 bit of memory, even in the best-performing models presented in this work, the per-neuron storage demand does not exceed 32 bits, demonstrating strong hardware feasibility. Moreover, the 1-bit storage overhead per spike highlights the potential for hardware deployment of SRC-SNN and ASRC-SNN with much longer skip spans.
\begin{table}[ht]
\centering
\caption{Computational metrics of SRC-SNN and ASRC-SNN on PS-MNIST}
\footnotesize 
\setlength{\tabcolsep}{4pt} 
\renewcommand{\arraystretch}{0.9} 
\begin{tabular}{cccccc}
\toprule
\multirow{2}{*}{\textbf{Model}} & \multirow{2}{*}{\textbf{$\lambda$/$T_\lambda$}} & \textbf{Memory} & \textbf{Training Time} & \textbf{Accuracy} \\
&        & \textbf{(GB)}   & \textbf{(hours)}        & \textbf{(\%)}     \\
\midrule
\multirow{5}{*}{SRC-SNN}
& 1 (vanilla) & 2.43  & 27.93 & 84.59 \\
& 2           & 2.67  & 28.23 & 90.65 \\
& 8           & 4.11  & 28.10 & 93.83 \\
& 16          & 6.39  & 28.20 & 94.48 \\
& 24          & 8.21  & 28.36 & 94.44 \\
\midrule
\multirow{5}{*}{ASRC-SNN}
& 11  & 5.12  & 32.86 & 95.15 \\
& 21  & 8.99  & 32.96 & 95.23 \\
& 31  & 10.21 & 32.90 & 95.22 \\
& 41  & 13.67 & 33.61 & 95.36 \\
& 51  & 15.20 & 33.43 & 95.40 \\
\bottomrule
\end{tabular}
\label{tab:training overhead}
\end{table}

\section{Some information related to the experiments}
Table.\ref{tab:convergence in ASRC} presents the final convergence values of the skip connection coefficients across different layers of ASRC-SNN as $T_{\lambda}$ increases.
\begin{table}[ht]
\centering
\caption{The skip coefficients for each layer of ASRC-SNN after training}
\footnotesize 
\setlength{\tabcolsep}{4pt} 
\renewcommand{\arraystretch}{0.9} 
\begin{tabular}{c c c c}
\toprule
\multicolumn{2}{c}{\textbf{PS-MNIST}} & \multicolumn{2}{c}{\textbf{SSC}} \\
\cmidrule(r){1-2} \cmidrule(l){3-4}
\textbf{$T_{\lambda}$} & \textbf{Final Skip coefficients} & \textbf{$T_{\lambda}$} & \textbf{Final Skip coefficients} \\
\midrule
6  & [5, 6, 3] & 6  & [2, 6, 6] \\
11  & [9, 10, 4] &11  & [2, 7, 8] \\
21  & [8, 12, 4] &21  & [3, 8, 13] \\
31  & [8, 14, 4] &31  & [3, 9, 19] \\
41  & [8, 13, 17] &41 & [3, 10, 23] \\
51  & [9, 25, 11] &51 & [3, 10, 27] \\
\bottomrule
\end{tabular}
\label{tab:convergence in ASRC}
\end{table}
\par
\begin{figure*}[h]
    \centering
    \begin{subfigure}[b]{\textwidth}
        \centering
        \includegraphics[width=0.8\textwidth]{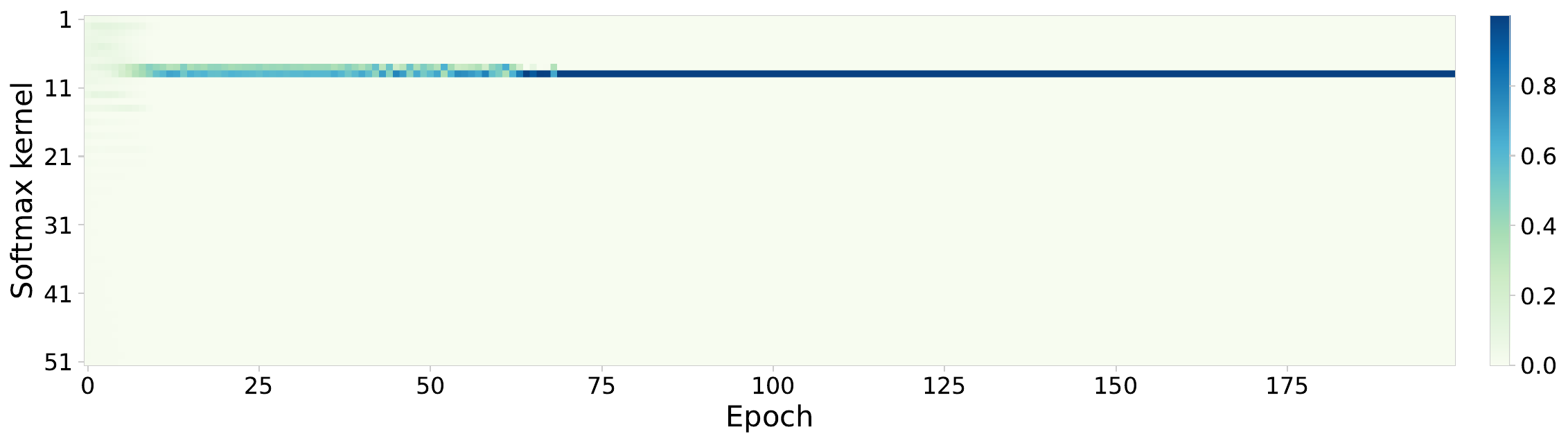}
        \caption{First layer}
        \label{fig:layer0}
    \end{subfigure}
    \hfill
    \begin{subfigure}[b]{\textwidth}
        \centering
        \includegraphics[width=0.8\textwidth]{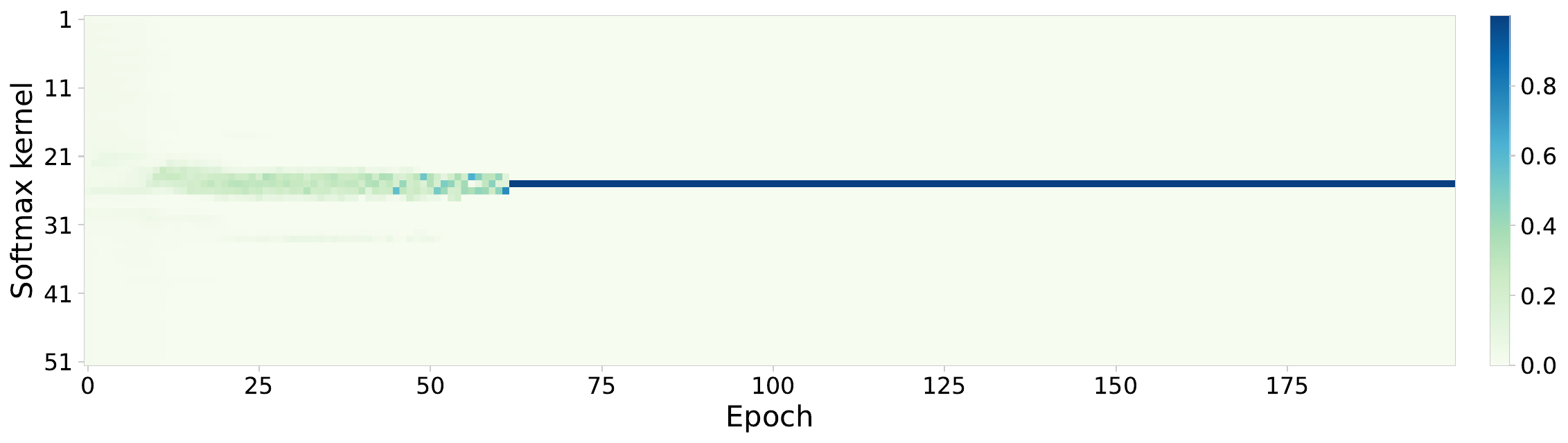}
        \caption{Second layer}
        \label{fig:layer1}
    \end{subfigure}
    \hfill
    \begin{subfigure}[b]{\textwidth}
        \centering
        \includegraphics[width=0.8\textwidth]{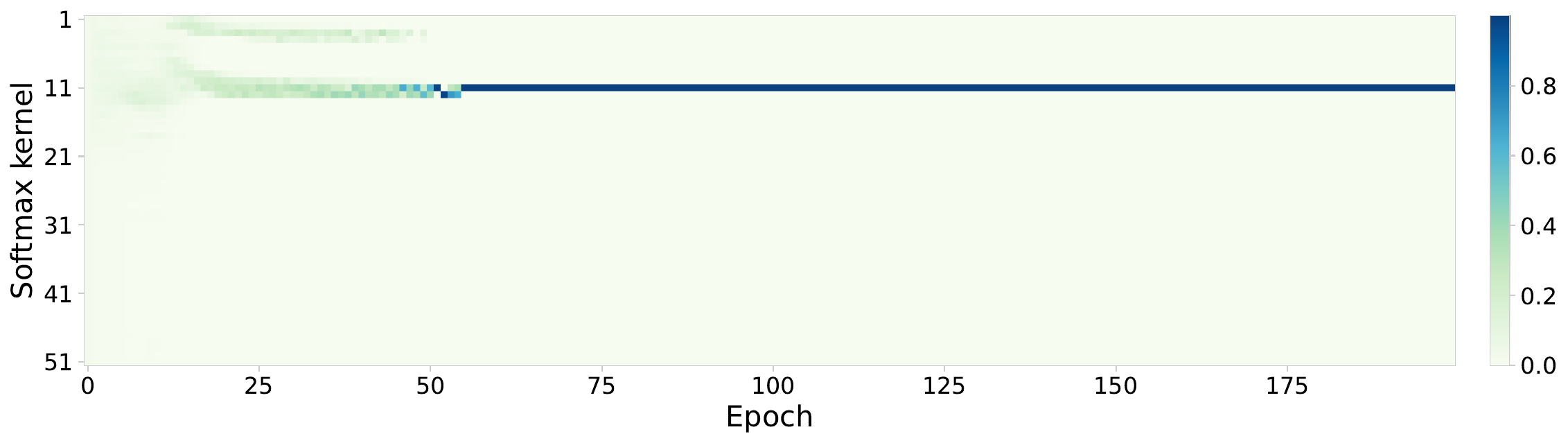}
        \caption{Third layer}
        \label{fig:layer2}
    \end{subfigure}
    \caption{These plots present heatmaps of the weight distributions of the Softmax kernels across different layers during the training of ASRC-SNN. The x-axis represents the epochs, the y-axis represents time, and each kernel has a size of $T_{\lambda}=51$. (a), (b), and (c) represent the first, second, and third layers, respectively.}
    \label{fig:heatmaps}  
\end{figure*}
\begin{figure*}[h]
    \centering
    \includegraphics[width=0.6\textwidth]{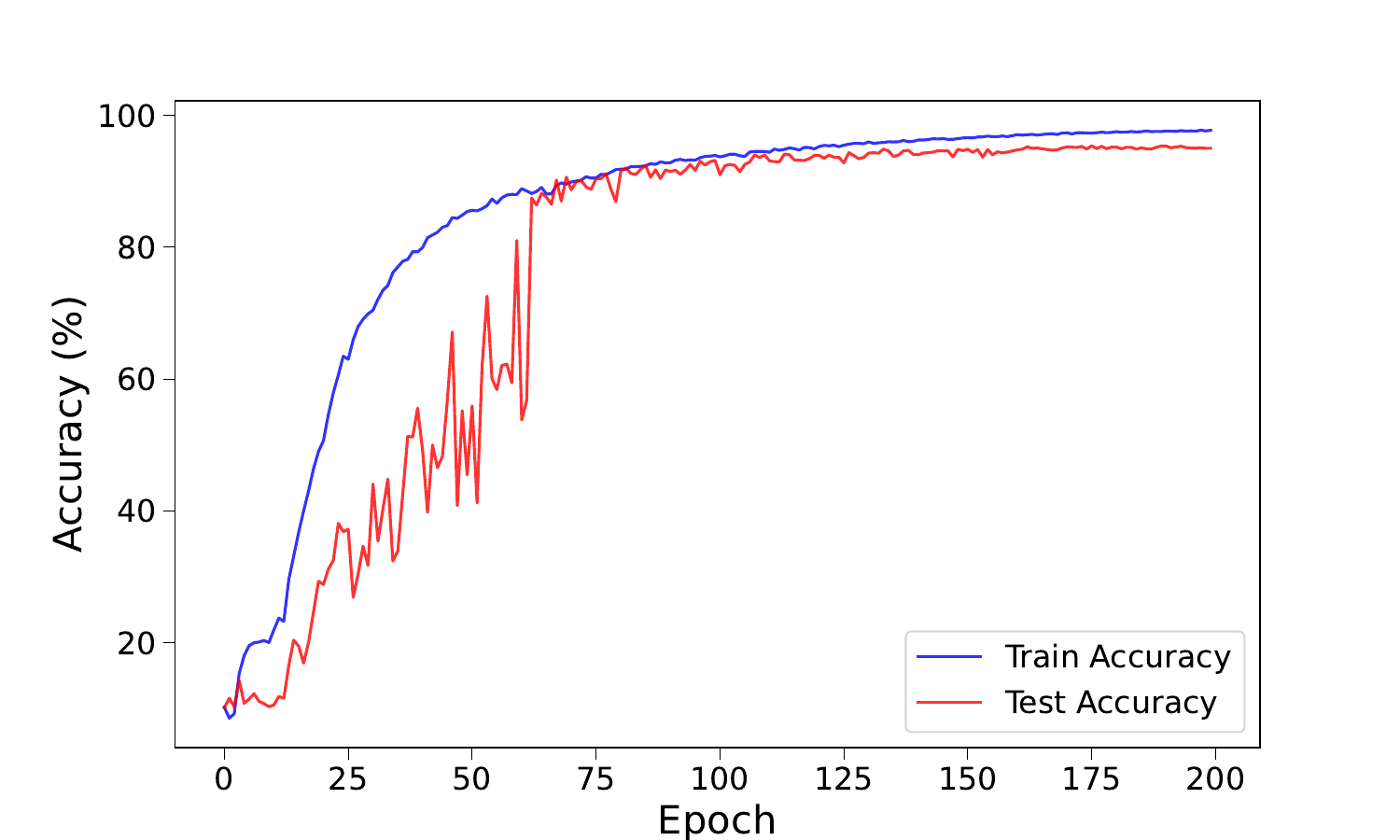}
    \caption{The accuracy change plot during the training of ASRC-SNN.}
    \label{fig:accuracy}
\end{figure*}

Figure.\ref{fig:heatmaps} presents heatmaps of the weight variations of the Softmax kernels across different layers during ASRC-SNN training on PS-MNIST. Figure.\ref{fig:accuracy} shows the corresponding accuracy change plot.
 
\section{other experiments}
\begin{table}[ht]
\centering
\caption{Performance of a variant of ASRC on PS-MNIST and SSC}
\footnotesize 
\setlength{\tabcolsep}{4pt} 
\renewcommand{\arraystretch}{0.9} 
\begin{tabular}{c c c c c c}
\toprule
\multicolumn{3}{c}{\textbf{PS-MNIST}} & \multicolumn{3}{c}{\textbf{SSC}} \\
\cmidrule(r){1-3} \cmidrule(l){4-6}
\textbf{$T_{\lambda}$} & \textbf{Hidden Size} & \textbf{Accuracy (\%)} & \textbf{$T_{\lambda}$} & \textbf{Hidden Size} & \textbf{Accuracy (\%)} \\
\midrule
6  & [64, 200, 200] & 95.99 & 6  & [256,256,256] & 81.35 \\
11  & [64, 190, 190] & 96.47 &11  & [253,253,253] & 81.19 \\
21  & [64, 170, 170] & 96.51 &21  & [250,250,250]& 80.99 \\
31  & [64, 150, 150] & 96.27 &31  & [248,248,248]& 81.15 \\
41  & [64, 120, 120] & 96.10 &41 & [245,245,245]& 80.73 \\
51  & [64, 90, 90] & 96.15 &51 & [243,243,243]& 80.80 \\
\bottomrule
\end{tabular}
\label{tab:a variant of ASRC}
\end{table}

\subsection{A Variant of ASRC}
In ASRC, each layer only needs to adaptively select a single skip span, resulting in a total of 
"$T_{\lambda} \times \text{number of layers}$" trainable parameters for learning skip connections. We extend ASRC by allowing each time step to independently adapt its own skip span, which increases the number of trainable parameters to 
"$T_{\lambda} \times \text{number of layers} \times \text{time steps}$".
To ensure that the total number of parameters remains comparable to the original ASRC, we appropriately reduce the dimensionality of the hidden layers. The performance of this variant on PS-MNIST and SSC is presented in Table.\ref{tab:a variant of ASRC}. We observe that, in most cases, it does not outperform the original ASRC, and only exhibits marginal improvements in a few cases.

\subsection{A Simple Exploration of Parameter-Sharing for k-ASRC}
Parameter-Sharing refers to sharing $W_2^l$ across $k$ skip connections as shown in Eq.~\eqref{eq:12}. When $k = 1$, $k$-ASRC degenerates to the original ASRC; when $1 < k < T_{\lambda}$, we introduce $k$ temperature-scaled softmax kernels to adaptively select $k$ skip connections, and then use a regular softmax kernel of length $k$ to assign weights to these $k$ kernels; when $k = T_{\lambda}$, we use only a regular softmax kernel. We set $T_{\lambda} = 21$, and the performance of $k$-ASRC on PS-MNIST is shown in Table~\ref{tab:k-asrc-performance}. We find that the performance of $k$-ASRC is optimal when $k = 1$, and the performance difference for other values of $k$ is minimal.

\begin{table}[ht]
\centering
\caption{Performance of $k$-ASRC with varying $k$ on PS-MNIST}
\begin{tabular}{c c}
\toprule
\textbf{$k$} & \textbf{Accuracy (\%)} \\
\midrule
1   & 95.23 \\
5   & 94.55 \\
10  & 94.66 \\
15  & 94.59 \\
20  & 94.64 \\
21  & 94.61 \\
\bottomrule
\end{tabular}
\label{tab:k-asrc-performance}
\end{table}

\end{document}